\documentclass{ecai} 

\usepackage{latexsym}
\usepackage{amssymb}
\usepackage{amsmath}
\usepackage{amsthm}
\usepackage{booktabs}
\usepackage[shortlabels]{enumitem}
\usepackage{graphicx}
\usepackage{color}

\usepackage[hidelinks]{hyperref}
\usepackage{adjustbox}
\usepackage{footnote}
\usepackage{xcolor,colortbl}
\usepackage{subcaption}
\usepackage{multirow}
\usepackage{algorithm2e}
\RestyleAlgo{ruled}

\newcommand{\BibTeX}{B\kern-.05em{\sc i\kern-.025em b}\kern-.08em\TeX}

\begin{document}


\begin{frontmatter}


\paperid{732} 


\title{Interpretable Graph Neural Networks for Tabular Data}


\author[A]{\fnms{Amr}~\snm{Alkhatib}
}
\author[A]{\fnms{Sofiane}~\snm{Ennadir}
}
\author[A]{\fnms{Henrik}~\snm{Boström}
}
\author[A,B]{\fnms{Michalis}~\snm{Vazirgiannis}
} 

\address[A]{KTH Royal Institute of Technology\\
    Electrum 229, 164 40 Kista, Stockholm, Sweden}
\address[B]{DaSciM, LIX, École Polytechnique, Institut Polytechnique de Paris, France.}


\begin{abstract}
Data in tabular format is frequently occurring in real-world applications. Graph Neural Networks (GNNs) have recently been extended to effectively handle such data, allowing feature interactions to be captured through representation learning. However, these approaches essentially produce black-box models, in the form of deep neural networks, precluding users from following the logic behind the model predictions. We propose an approach, called IGNNet (Interpretable Graph Neural Network for tabular data), which constrains the learning algorithm to produce an interpretable model, where the model shows how the predictions are exactly computed from the original input features. A large-scale empirical investigation is presented, showing that IGNNet is performing on par with state-of-the-art machine-learning algorithms that target tabular data, including XGBoost, Random Forests, and TabNet. At the same time, the results show that the explanations obtained from IGNNet are aligned with the true Shapley values of the features without incurring any additional computational overhead.
\end{abstract}

\end{frontmatter}


\section{Introduction}

In some application domains, e.g., medicine and law, predictions made by machine learning models need justification for legal and ethical considerations \cite{LakkarajuKCL17,Goodman2017}. In addition, users may put trust in such models only with a proper understanding of the reasoning behind the predictions. A direct solution is to use learning algorithms that produce interpretable models, such as logistic regression \cite{log_reg}, which provides both local (instance-specific) and global (model-level) explanations for the predictions. However, such algorithms often result in a substantial loss in predictive performance compared to algorithms that generate black-box models, e.g., XGBoost \cite{xgboost}, Random Forests \cite{Breiman2001}, and deep learning algorithms \cite{Mori2019,a13010017}. Post-hoc explanation techniques, e.g., SHAP \cite{SHAP}, LIME \cite{LIME}, and Anchors \cite{Anchors}, have been put forward as tools  to explain predictions of the black-box models. However, the explanations provided by such methods are often computationally intensive \cite{alkhatib23a,jethani2022fastshap} and lack fidelity guarantees, i.e., there are no guarantees that the provided explanation accurately reflects the underlying model, as studied previously \cite{covert21a,delaunay:hal-03133223,NEURIPS2019_a7471fdc}. As extensively argued in \cite{rudin2019stop}, there are hence several reasons to consider generating interpretable models in the first place, if trustworthiness is a central concern. 

Graph Neural Networks (GNNs) have emerged as a powerful framework for representation learning of graph-structured data \cite{xu2018how}. The application of GNNs has been extended to tabular data, where a GNN can be used to learn an enhanced representation for the data points (rows) or to model the interaction between different features (columns). TabGNN \cite{TabGNN} is an example of the first approach, where each data point is represented as a node in a graph. In comparison, TabularNet \cite{du2021tabularnet} and Table2Graph \cite{ijcai2022p336} follow the second approach, where the first uses a Graph Convolutional Network to model the relationships between features, and the second learns a probability adjacency matrix for a unified graph that models the interaction between features of the data points. GNNs can also be combined with other algorithms suited for tabular data, e.g., as in BGNN \cite{ivanov2021boost}, which combines gradient-boosted decision trees and a GNN in one pipeline, where the GNN addresses the graph structure and the gradient-boosted decision trees handle the heterogeneous features of the tabular data. To the best of our knowledge, all previous approaches to using GNNs for tabular data result in black-box models and they are hence associated with the issues discussed above when applied in contexts with strong requirements on trustworthiness. In this work, we propose a novel GNN approach for tabular data, with the aim to eliminate the need to apply post-hoc explanation techniques without sacrificing predictive performance.

The main contributions of this study are:
\begin{itemize}
    \item a novel approach, called \textbf{I}nterpretable \textbf{G}raph \textbf{N}eural \textbf{Net}work for tabular data (IGNNet), that exploits powerful graph neural network models while still being able to show exactly how the prediction is derived from the input features in a transparent way

    \item a large-scale empirical investigation evaluating the explanations of IGNNet as well as comparing the predictive performance of IGNNet to state-of-the-art approaches for tabular data; XGBoost, Random Forests, and multi-layer perceptron (MLP), as well as to an algorithm generating interpretable models; TabNet \cite{Arik_Pfister_2021}
    
    \item an ablation study comparing the performance of the proposed approach to a black-box version, i.e., not constraining the learning algorithm to produce transparent models for the predictions
    
\end{itemize}

In the next section, we briefly review related work. In Section~\ref{IGNNet}, we describe the proposed interpretable graph neural network. In Section~\ref{Evaluation}, results from a large-scale empirical investigation are presented and discussed, in which the explanations of the proposed method are evaluated and the performance is compared both to interpretable and powerful black-box models. Section~\ref{limitations} discusses the limitations of the proposed approach. 
Finally, in the concluding remarks section, we summarize the main conclusions and point out directions for future work.

\section{Related Work}

In this section, we provide pointers to self-explaining (regular and graph) neural networks, and briefly discuss their relation to model-agnostic explanation techniques and interpretable models. We also provide pointers to work on interpretable deep learning approaches for tabular data.

\subsection{Self-Explaining Neural Networks}\label{related_NNs}

Several approaches to generating so-called self-explaining neural networks have been introduced in the literature; in addition to generating a prediction model, they all incorporate a component for explaining the predictions. They can be seen as model-specific explanation techniques, in contrast to model-agnostic techniques, such as LIME and SHAP, but sharing the same issues regarding fidelity and lack of detail regarding the exact computation of the predictions.

Approaches in this category include the method in \cite{lei-etal-2016-rationalizing}, which is an early self-explaining neural network for text classification, the Contextual Explanation Network (CEN) \cite{Shedivat}, which  generates explanations using intermediate graphical models, the Self-explaining Neural Network (SENN) \cite{NEURIPS2018_SENN}, which generalizes linear classifiers to neural networks using a concept autoencoder, and the CBM-AUC \cite{9758745_CBM_AUC}, which improves the efficiency of the former by replacing the decoder with a discriminator. Some approaches generate explanations in the form of counterfactual examples, e.g., CounterNet \cite{guo2021counternet}, and Variational Counter Net (VCNet) \cite{Guyomard2022VCNetAS}. Again, such explanations do not provide detailed information on how the original predictions are computed and how exactly the input features affect the outcome.

\subsection{Self-Explaining Graph Neural Networks}\label{related_GNNs}

The Self-Explaining GNN (SE-GNN) \cite{SEGNN/Enyan.3482306} uses similarities between nodes to make predictions on the nodes' labels and provide explanations using the most similar K nodes with labels. 
ProtGNN \cite{Zhang2022ProtGNNTS} also computes similarities, but between the input graph and prototypical graph patterns that are learned per class. Cui et al. \cite{Interpretable_GNN_Cui} proposed a framework to build interpretable GNNs for connectome-based brain disorder analysis that resembles the signal correlation between different brain areas. Feng et al. \cite{KerGNNs} proposed the KerGNN (Kernel Graph Neural Network), which improves model interpretability using graph filters. The graph filters learned in KerGNNs can uncover local graph structures in a dataset. Xuanyuan et al. \cite{Xuanyuan_} proposed scrutinizing individual neurons in a GNN to generate global explanations using neuron-level concepts. CLARUS \cite{METSCH2024104600} enhances human understanding of GNN predictions in medicine and facilitates the visualization of patient-specific information.

The self-explainable GNN methods mentioned above are not suited for tabular data, as they are designed for input data that are inherently graphical, such as social networks and molecular structures.

\subsection{Interpretable Deep Learning for Tabular Data}\label{related_interpretable}

In an endeavor to provide an interpretable regression model for tabular data while retaining the performance of deep learning models, and inspired by generalized linear models (GLM), LocalGLMnet was proposed to make the regression parameters of a GLM feature dependent, allowing for quantifying variable importance and also conducting variable selection \cite{LocalGLMnet}. TabNet \cite{Arik_Pfister_2021} is another interpretable method proposed for tabular data learning, which employs a sequential attention mechanism and learnable masks for selecting a subset of meaningful features to reason from at each decision step. The feature selection is instance-based, i.e., it differs from one instance to another. The feature selection masks can be visualized to highlight important features and show how they are combined. However, it is not obvious how the features are actually used to form the predictions. 

\section{The Proposed Approach: IGNNet}
\label{IGNNet}

This section describes the proposed method to produce an interpretable model using a graph neural network. We first outline the details of a GNN for graph classification and then show how it can be constrained to produce interpretable models. Afterward, we show how it can be applied to tabular data. Finally, we show how the proposed approach can maintain both interpretability and high performance.

\subsection{Interpretable Graph Neural Network}
\label{sec:gnn}

The input to a GNN learning algorithm is a set of graphs denoted by $\mathcal{G} = (V, E, X,  \textit{\textbf{A}})$, consisting of a set of nodes $V$, a set of edges $E$, a set of node feature vectors $X$, and an adjacency matrix $\textit{\textbf{A}}$, where $V = \{v_1, \ldots, v_N\}$, $E \subseteq \{(v_i, v_j) | v_i, v_j \in V\}$, $X = \{\textbf{x}_1, \ldots, \textbf{x}_N\}$, and the weight of edge $(v_i, v_j)$ is represented by a scalar value $\delta_{i,j}$ in the weighted adjacency matrix $\textit{\textbf{A}}$, where $\delta_{i,j} = \textit{\textbf{A}}(i,j)$. 
A GNN algorithm learns a representation vector $\textbf{h}_i$ for each node $v_i$, which is initialized as $\textbf{h}_i^{(0)} = \textbf{x}_i$. The key steps in a GNN for graph classification can be summarized by the following two phases \cite{xu2018how}:

\begin{enumerate}[(a)]
    \item \textbf{Message Passing:} Each node passes a message to the neighboring nodes, then aggregates the passed information from the neighbors. Finally, the node representation is updated with the aggregated information. A neural network can also be used to learn some message functions between nodes. The message passing phase can be formulated as:

    \begin{equation}
    \label{eq_wsum}
    \textbf{h}_i^{(l+1)} = \varphi\left(\textbf{w}^{(l)}\left(\delta_{i,i} \textbf{h}_i^{(l)} + \sum_{u \in \mathcal{N}(i)} \delta_{i,u} \textbf{h}_u^{(l)}\right)\right)
    \end{equation}
    
    Where $\textbf{h}_i^{(l)}$ is the hidden representation of the node $v_i$ in the $l$-th layer, $\mathcal{N}(i)$ is the neighborhood of node $v_i$, $\textbf{w}^{(l)}$ represents the learnable parameters, and $\varphi$ is a non-linearity function. 
    
    The adjacency matrix $\textit{\textbf{A}}$ of size $|V| \times |V|$ contains the edge weights and can be normalized similar to a Graph Convolutional Network (GCN) \cite{kipf2017semisupervised} as shown in \eqref{eq_adj_norm}.

        \begin{equation}
        \label{eq_adj_norm}
        \Tilde{\textit{\textbf{A}}} = \textit{\textbf{D}}^{-\frac{1}{2}} \textit{\textbf{A}} \textit{\textbf{D}}^{-\frac{1}{2}}
        \end{equation}
    Here $\textit{\textbf{D}}$ is the degree matrix $\textit{\textbf{D}}_{ii} = \sum_{j} \textit{\textbf{A}}_{ij}$ \cite{kipf2017semisupervised}.
    \item \textbf{Graph Pooling (Readout):} A representation of the whole graph $\mathcal{G}$ is learned using a simple or advanced function \cite{xu2018how}, e.g, sum, mean, or MLP. 
    
\end{enumerate}
The whole graph representation obtained from the \textbf{graph pooling} phase can be submitted to a classifier to predict the class of the graph, which can be trained in an end-to-end architecture \cite{end_to_end_GC,Hier_GRL}.

\textit{The pooling function can be designed to provide an interpretable graph classification layer}. Thus, the final hidden representation of each node is mapped to a single  value, for instance, through a neural network layer or dot product  ($\mathcal{R}(\textbf{h}_i^{(l+1)} \in \mathbb{R}^n) = h_i \in \mathbb{R}^1$), and concatenated to obtain the final representation $\textbf{g}$ of the graph $\mathcal{G}$ where a scalar value in $\textbf{g}$ corresponds to a node in the graph. Consequently, if a set of weights is applied to classify the graph, we can trace the contribution of each node to the predicted outcome, i.e., the user can find out which nodes contributed to the predicted class. For example, $\textbf{g}$ can be used directly as follows:

\begin{equation}
    \hat{y} = \text{link} \left(\sum_{i=1}^n{w_i g_i}\right)
\label{eq5}
\end{equation}

\noindent where $w_i$ is the weight (in vector $\textbf{w}$) assigned to node $v_i$ represented in $g_i$.
The link function is applied to accommodate a valid range of outputs, e.g., the sigmoid function for binary and softmax for multi-class classification. This is equivalent to:

\begin{equation}
    \hat{y} = \text{link} \left(\sum_{i=1}^n{w_i \mathcal{R}(\textbf{h}_i^{(l+1)})}\right)
\label{eq_6}
\end{equation}

In the case of binary classification, one vector of weights ($\textbf{w}$) is applied, and for multiple classes, each class has a separate vector of weights.

\subsection{Representing Tabular Data Points as Graphs}

\begin{figure*}[ht]
    \centering
    \includegraphics[width=.9\textwidth]{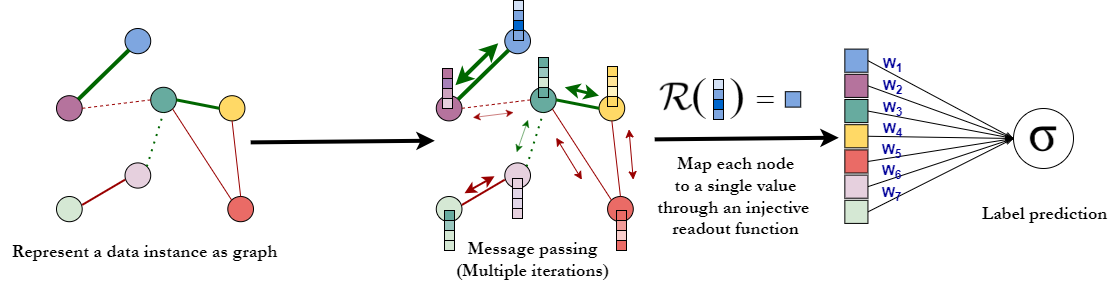}
    \vspace{4 mm}
    \caption{\textbf{An overview of our proposed approach.} Each data instance is represented as a graph by embedding the feature values into a higher dimensionality, and the edge between two features (nodes) is the correlation value. Multiple iterations of message passing are then applied. Finally, the learned node representation is projected into a single value, and a whole graph representation is obtained by concatenating the projected values.}
    \label{fig:illust1}
    \vspace{4 mm}
\end{figure*}

The proposed readout function in the previous subsection allows for determining the contribution of each node in a prediction, if a white-box classification layer is used for the latter. Therefore, we propose representing each data instance as a graph where \textit{the features are the nodes} of that graph and \textit{the linear correlation between features are the edge weights}, as we assume that not all features are completely independent. The initial representation of a node is a vector of one dimension, and the value is just the feature value, which can be embedded into a higher dimensionality. The idea is outlined in Algorithm~\ref{alg:1} and illustrated in Figure \ref{fig:illust1}.

\begin{algorithm}
\SetAlgoLined
\KwData{a set of graphs $\mathbb{G}$ and labels $\mathbb{Y}$}
\KwResult{Model parameters $\theta$}
Initialize $\theta$\\
\For{number of training iterations}{
$\mathcal{L} \gets 0$\\
\For{each $\mathcal{G}_j \in \mathbb{G}$}{
$\textit{\textbf{H}}^{(0)}_j \gets \mathcal{G}_j$\\
\For{each layer l $\in$ messagePassing layers}{
        $\textit{\textbf{H}}^{(l+1)}_j \gets $messagePassing$(\textit{\textbf{H}}^{(l)}_j)$\\
        }
$\textbf{g}_j \gets $readout$(\textit{\textbf{H}}^{(l+1)}_j)$\\ 
$\hat{y}_j \gets $predict$(\textbf{g}_j)$\\
$\mathcal{L} \gets \mathcal{L} + $loss$(\hat{y}_j,y_j \in \mathbb{Y})$
}
Compute gradients $\nabla_{\theta} \mathcal{L}$\\
Update $\theta \gets \theta - \nabla_{\theta} \mathcal{L}$
}
\caption{IGNNet}
\label{alg:1}
\end{algorithm}

In order to make a prediction for a test instance, the data point has to be converted to a graph using the same procedure for building the input graphs to IGNNet, and for which graph node representations are obtained using a GNN with parameters $\theta$. Finally, the output layer is used to form the prediction.

\subsection{How can IGNNet achieve high performance while maintaining interpretability?}

An expressive GNN can potentially capture complex patterns and dependencies in the graph, allowing nodes to be mapped to distinct representations based on their characteristics and relationships \cite{GNNBook_ch5}. 
Moreover, a GNN with an injective aggregation scheme can not only distinguish different structures but also map similar structures to similar representations \cite{xu2018how}. Therefore, if the tabular data are properly presented as graphs, GNNs with the aforementioned expressive capacities can model relationships and interactions between features, and consequently approximate complex non-linear mappings from inputs to predictions. On top of that, it has been shown by \cite{ennadir2023a} that GCNs based on 1-Lipschitz continuous activation functions can be improved in stability and robustness with Lipschitz normalization and continuity analysis; similar findings have also been demonstrated on graph attention networks (GAT) \cite{dasoulas21a}. This property is of particular importance when the application domain endures adversarial attacks or incomplete tabular data.

The proposed readout function in subsection \ref{sec:gnn} can produce an interpretable output layer. However, it does not guarantee the interpretability of the whole GNN without message-passing layers that consistently maintain relevant representations of the input features. Accordingly, we constrain the message-passing layer to produce interpretable models using the following conditions:

\begin{enumerate}
    \item Each feature is represented in a distinct node throughout the consecutive layers.

    \item Each node has a highly weighted self-loop which conveys the main message of the node.
    
    \item Each node is bounded to interact with a particular neighborhood, where it maintains correlations with the nodes within that neighborhood. 
\end{enumerate}

The message-passing operation is based on the linear relationships between nodes. The strength of a passed message is directly proportional to the linear relationship between the two nodes, and the sign of the correlation value determines the signs of the messages. In addition, the highly weighted self-loops maintain the main messages carried by the nodes and prevent them from fading away through multiple message-passing layers. As a result, the aggregated messages could potentially hold significance to the input feature values.
The proposed graph pooling function, combined with the constrained message-passing layers that keep representative information about the input features, allows tracking each feature's contribution at the output layer and also through the message-passing layers all the way to the input features.

\section{Empirical Investigation}
\label{Evaluation}

This section evaluates both the explanations and predictive performance of IGNNet. We begin by outlining the experimental setup, then by evaluating the explanations produced by IGNNet, and lastly, we benchmark the predictive performance.

\subsection{Experimental Setup} \label{experimental_setup}

\begin{figure*}[ht]
    \centering
    \includegraphics[width=.9\textwidth]{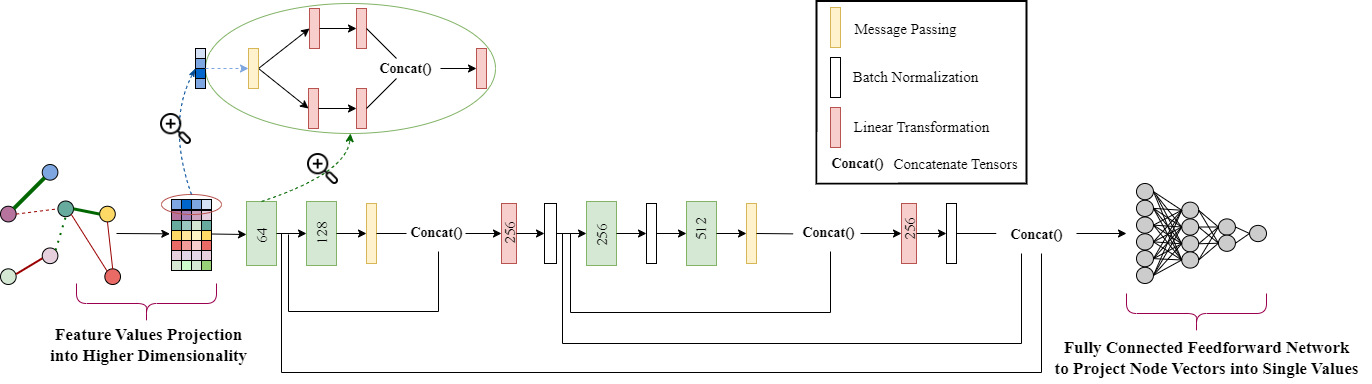}
    \vspace{4 mm}
    \caption{\textbf{IGNNet default architecture.} It starts with a layer to project the features into higher dimensionality, a linear transformation from one dimension to 64 dimensions. A Relu activation function follows each message-passing layer and each green block as well. The feedforward network at the end has no activation functions between layers to ensure a linear transformation into a single value. A sigmoid activation function follows the feedforward network to obtain the final value for each feature between 0 and 1.}
    \label{fig:illust3}
    \vspace{4 mm}
\end{figure*}

A GNN consists of one or more message-passing layers, and each layer can have a different design, e.g., different activation functions and batch normalization, which is the intra-layer design level \cite{GNN_Design_Space}. There is also the inter-layer design level, which involves, for instance, how the message-passing layers are organized into a neural network and if skip connections are added between layers \cite{GNN_Design_Space}. While the intra-layer and inter-layer designs can vary based on the nature of the prediction task, we propose a general architecture for our empirical investigation.\footnote{The source code is available here: \url{https://github.com/amrmalkhatib/ignnet}} However, it is up to the user to modify the architecture of the GNN. The number of message-passing layers, number of units in linear transformations, and other hyperparameters were found based on a quasi-random search and evaluation on development sets of the following three datasets: Churn, Electricity, and Higgs. We have six message-passing layers in the proposed architecture, each with a Relu activation function. Multiple learnable weights are also applied to the nodes' representation, followed by a Relu function. Besides three batch normalization layers, four skip connections are added as illustrated in Figure \ref{fig:illust3}. After all the GNN layers, we use a feedforward neural network (FNN) to map the multidimensional representation of each node into a single value. In the FNN, we do not include any activation functions in order to keep the mapping linear, but a sigmoid function is applied after the final layer to obtain a value between 0 and 1 for each node. The FNN is composed of 8 layers with the following numbers of units (128, 64, 32, 16, 8, 4, 2, 1) and 3 batch normalization layers after the second, fourth, and sixth hidden layers. After the FNN, the nodes' final values are concatenated to form a representation of the whole graph (data instance). Finally, the weights that are output are used to make predictions. The GNN is trained end-to-end, starting from the embeddings layer and ending with the class prediction.

We also provide an opaque variant of IGNNet (OGNNet, opaque graph neural net for tabular data), where the FNN, along with the output layer, is replaced by an MLP of one hidden layer with 1024 hidden units and a Relu activation function. All the learned node representations just before the FNN are concatenated and passed to the MLP for class prediction. The OGNNet is introduced to determine, by an ablation study, how much predictive performance we may lose by squashing the learned multidimensional representation of the nodes into scalar values and applying a white box classifier instead of a black box.

In the experiments, 35 publicly available datasets are used.\footnote{All the datasets were obtained from \url{https://openml.org}} Each dataset is split into training, development, and test sets. The development set is used for overfitting detection and early stopping of the training process, the training set is used to train the model, and the test set is used to evaluate the model.\footnote{Detailed information about each dataset is provided in the appendix} For a fair comparison, all the compared learning algorithms are trained without hyperparameters tuning using the default settings on each dataset. In cases where the learning algorithm does not employ the development set to measure performance progress for early stopping, the development and training subsets are combined into a joint training set. The adjacency matrix uses the correlation values computed on the training data split. The weight on edge from the node to itself (self-loop) is a user-adjustable hyperparameter, constitutes between 70\% to 90\% (on average) of the weighted summation to keep a strong message per node that does not fade out with multiple layers of message-passing. Weak correlation values are excluded from the graph, so if the absolute correlation value is below 0.2, the edge is removed unless no correlation values are above 0.2; in case of the latter, the procedure is repeated using a reduced threshold of 0.05.\footnote{The appendix includes an ablation study on how various hyperparameter preferences affect predictive performance.}
The Pearson correlation coefficient \cite{Pearson_corr} is used to estimate the linear relationship between features. In the data preprocessing step, the categorical features are binarized using one-hot encoding, and all the feature values are normalized using min-max normalization (the max and min values are computed on the training split). The normalization keeps the feature values between 0 and 1, which is essential for the IGNNet to have one scale for all nodes.

The following algorithms are also evaluated in the experiments: XGBoost, Random Forests, MLP and TabNet. XGBoost and Random Forests are trained on the combined training and development sets. The MLP has two layers of 1024 units with Relu activation function and is trained on the combined training and development sets with early stopping and 0.1 validation fraction. TabNet is trained with early stopping after 20 consecutive epochs without improvement on the development set, and the best model is used in the evaluation.

For imbalanced binary classification datasets, we randomly oversample the minority class in the training set to align the size with the majority class. All the compared algorithms are trained using the oversampled training data. While for multi-class datasets, no oversampling is conducted.  

The area under the ROC curve (AUC) is used to measure the predictive performance, as it is not affected by the decision threshold. For the multi-class datasets, weighted AUC is calculated, i.e., the AUC is computed for each class against the rest and weighted by the support. 

\begin{figure*}[ht]

\centering
\includegraphics[width=.34\textwidth]{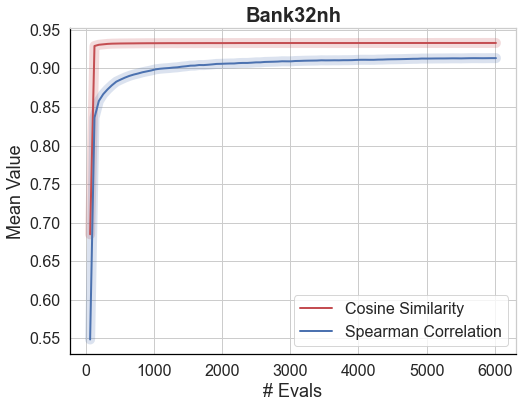}\hfill
\includegraphics[width=.32\textwidth]{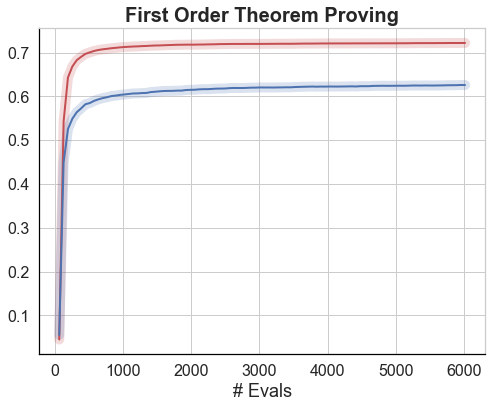}\hfill
\includegraphics[width=.32\textwidth]{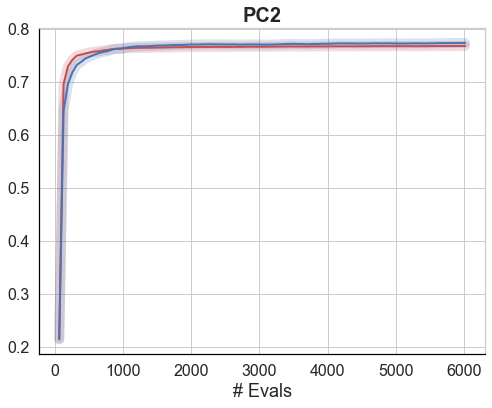}
\vspace{4 mm}
\caption{\textbf{Comparison of KernelSHAP's approximations and the importance scores obtained from IGNNet.} We measure the similarity of KernelSHAP's approximations to the scores of IGNNet at each iteration of data sampling and evaluation of KernelSHAP. KernelSHAP exhibits improvement in approximating the scores derived from IGNNet with more data sampling.}
\label{fig:transparency}
\vspace{4 mm}
\end{figure*}

\subsection{Evaluation of Explanations}

The feature scores produced by IGNNet should ideally reflect the contribution of each feature toward the predicted outcome and, therefore, they should be equivalent to the true Shapley values.
As it has been shown that KernelSHAP converges to the true Shapley values when provided with an infinite number of samples \cite{covert21a,jethani2022fastshap}, it is anticipated that the explanations generated by KernelSHAP will progressively converge to more similar values to the scores of IGNNet as the sampling process continues. This convergence arises from KernelSHAP moving towards the true values, while the scores of IGNNet are expected to align with these true values. To examine this conjecture, we explain IGNNet using KernelSHAP and measure the similarity between KernelSHAP's explanations and IGNNet's scores following each iteration of data sampling and KernelSHAP evaluation. For the feasibility of the experiment, 500 examples are randomly selected from the test set of each dataset to be explained. The cosine similarity and Spearman rank-order correlation are used to quantify the similarity between explanations. The cosine similarity measures the similarity in the orientation, while the Spearman rank-order measures the similarity in ranking the importance scores \cite{Amir_02488}.

The results demonstrate a general trend wherein KernelSHAP's explanations converge to more similar values to IGNNet's scores across various data instances and the 35 datasets, as depicted in Figure \ref{fig:transparency}. The consistent convergence to more similar values clearly indicates that IGNNet provides transparent models with feature scores aligned with the true Shapley values.\footnote{The complete results of the 35 datasets are provided in the appendix.}

\begin{figure*}
  \centering
  \subfloat[The original data point.]{\includegraphics[width=0.48\textwidth]{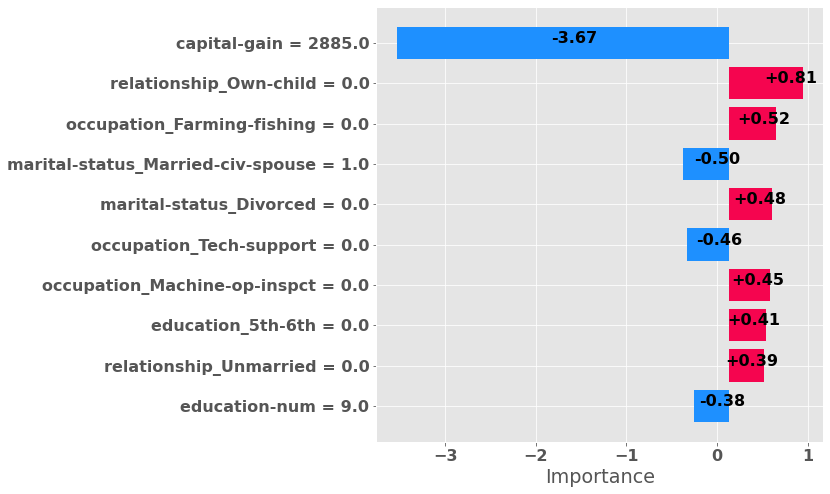}\label{fig:adult_gnn}}
  \hfill
  \subfloat[The data point with a modified capital gain value. ]{\includegraphics[width=0.48\textwidth]{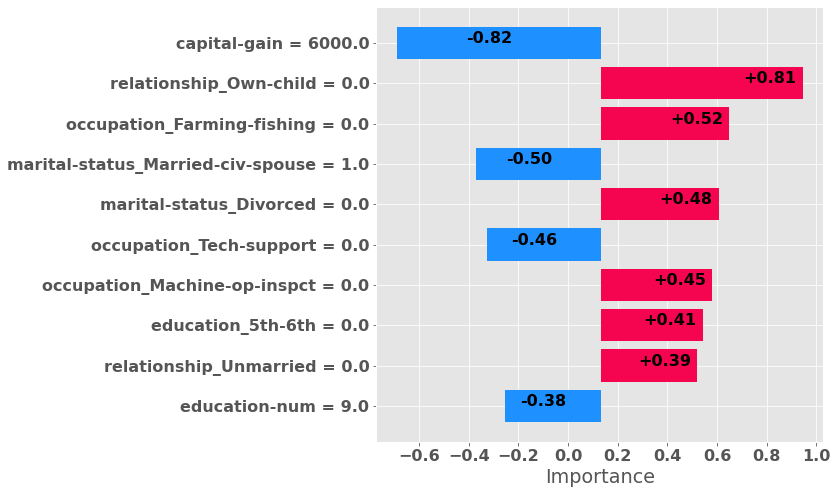}\label{fig:adult_mod_gnn}}
  \vspace{5 mm}
  \caption{Explanation to a single prediction on Adult dataset.}
  \vspace{4 mm}
  \label{fig:post_hoc}
\end{figure*}

\subsection{Illustration of Explanations}\label{exp_illustration}

In this section, we show, using a toy example, how the computed feature scores by IGNNet can be used to understand the feature contributions toward a specific prediction. Note that this is done in exactly the same way as how one would interpret the predictions of a logistic regression model or the feature importance scores generated by the SHAP \citep{SHAP} or LIME \citep{LIME} explainer. As the computed feature scores reveal exactly how IGNNet formed the prediction, the user can directly see which features have the greatest impact on the final prediction, and possibly also how they may be modified to affect the outcome. To demonstrate this, we present the feature scores for predictions made by IGNNet using an example from the Adult dataset.\footnote{We provide another example from the Churn dataset in the appendix.} In the following illustration, we display the feature scores centered around the bias value, which, when summed with the bias, will produce the exact outcome of IGNNet if the sigmoid function is applied. The scores are sorted according to their absolute values, and only the top 10 features are plotted for ease of presentation. A displayed score $\tau_i$ of feature $x_i$ represents all the weights and the computations applied to the input value, as shown in equation~\ref{eq_illust}.

\begin{equation}
\begin{aligned}
\tau_i =& w_{i} \mathcal{R}(\varphi(\textbf{w}^{(l)}(\sum_{u \in \mathcal{N}(i)} \delta_{i,u} \textbf{h}_u^{(l)} \\
&+ \delta_{i,i} \varphi(\textbf{w}^{(l-1)}(\sum_{u \in \mathcal{N}(i)} \delta_{i,u} \textbf{h}_u^{(l-1)} \\
&+ \delta_{i,i} (...\varphi(\textbf{w}^{(0)}(\sum_{u \in \mathcal{N}(i)} \delta_{i,u} \textbf{x}_u + \delta_{i,i} \textbf{x}_i)...))))))
\end{aligned}
\label{eq_illust}
\end{equation}

IGNNet predicted the negative class ($\leq$ 50K) with a narrow margin (0.495). The explanation shows that a single feature (capital-gain=2885) has the highest contribution compared to any other feature value. In the training data,  the capital-gain has a maximum value of 99999.0, a minimum value of 0, a mean value of 1068.36, and a 7423.08 standard deviation. Therefore, the capital-gain is relatively low for the selected data instance. To test if the explanation reflects the actual reasoning of IGNNet, we raise the capital-gain value by a smaller value than the standard deviation to be 6000 while leaving the remaining feature values constant, and it turns out to be enough to alter the prediction to a positive ($>$ 50K) with 0.944 as the predicted value. We can also see that the negative score of the capital-gain feature went from -3.67 in the original instance (\ref{fig:adult_gnn}) to -0.82 in the modified instance, as shown in Figure \ref{fig:adult_mod_gnn}. So the user can adjust the value of an important feature as much as needed to alter the prediction.

\subsection{Evaluation of Predictive Performance}\label{baseline}

\begin{figure*}[ht]
    \centering
    \includegraphics[width=.8\textwidth]{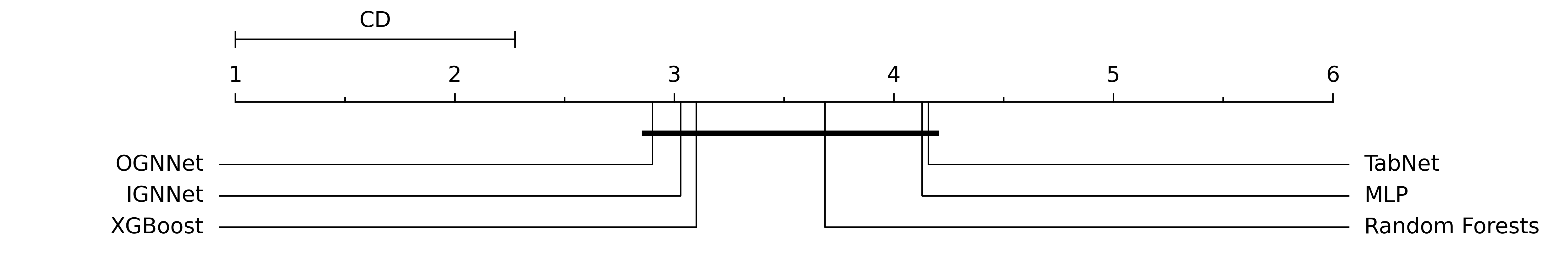}
    \vspace{1 mm}
    \caption{\textbf{The average rank of the compared classifiers on the 35 datasets with respect to the AUC} (a lower rank is better), where the critical difference (CD) represents the largest difference that is not statistically significant.}
    \label{fig:ranks}
    \vspace{5 mm}
\end{figure*}

Detailed results for IGNNet and the five competing algorithms on the 35 datasets are shown in Table~\ref{table:experiments}. The ranking of the six algorithms across 35 datasets, based on their AUC values, reveals OGNNet to exhibit superior performance, claiming the top position, closely followed by IGNNet and XGBoost. In order to investigate whether the observed differences are statistically significant, the Friedman test \cite{Friedman_test} was employed, which indeed allowed to reject the null hypothesis, i.e., reject that there is no difference in predictive performance, as measured by the AUC, at the 0.05 level. The result of subsequently applying the post-hoc Nemenyi test \cite{nemenyi:distribution-free} to determine what pairwise differences are significant, again at the 0.05 level, is summarized in Figure \ref{fig:ranks}. However, the result shows no specific significant pairwise differences between any of the compared algorithms. Furthermore, the results show that using IGNNet instead of the black-box variant, OGNNet, does not significantly reduce the predictive performance while maintaining performance at the level of other powerful algorithms for tabular data, e.g., XGBoost and Random Forests.

\subsection{Computational Cost}\label{cost}
 
The computational cost varies based on the architecture of the graph neural network, which can be altered and determined based on the predictive task and the dataset. The cost relative to the conventional graph neural networks remains the same, i.e., the cost does not increase with the proposed approach. The computational cost also depends on the number of features in the dataset. Therefore, the user can decide on the suitable architecture and the acceptable computational cost.

\section{Limitations}
\label{limitations}

The proposed approach targets tabular data primarily consisting of numerical features. However, datasets with substantial categorical features impose two challenges. Firstly, Pearson correlation is inadequate in handling correlation between categorical and numerical features or between two categorical features. Secondly, we use one-hot encoding for categorical features, which can result in an exponential growth in dimensionality when dealing with nominal features with numerous categories. Another limitation arises when input features lack correlation, resulting in a null graph (a completely disconnected graph).

\begin{table*}[ht]
\caption{The AUC of IGNNet, OGNNet, MLP, Random Forests, and XGBoost. The best-performing model is \colorbox[HTML]{A8B6E0}{colored in blue}, and the second best-performing is \colorbox[HTML]{E0EAF8}{colored in light blue}.}
\vspace{2 mm}
\centering
\begin{adjustbox}{width=.68\textwidth}
\small
\begin{tabular}{l c c c c c c}
    \toprule
    \rowcolor[HTML]{EFEFEF} 
    
    \multicolumn{1}{c}{{\cellcolor[HTML]{EFEFEF}Dataset}} & \multicolumn{1}{l}{\cellcolor[HTML]{EFEFEF}IGNNet} & OGNNet     & TabNet & MLP    & Random Forests   & XGBoost    \\
    \cmidrule(lr){1-7}
    Abalone & \colorbox[HTML]{E0EAF8}{0.881} & \colorbox[HTML]{A8B6E0}{0.883} & 0.857 & 0.877 & 0.876 & 0.869\\ 
    Ada Prior & \colorbox[HTML]{A8B6E0}{0.905} & 0.888 & 0.848 & 0.877 & 0.885 & \colorbox[HTML]{E0EAF8}{0.894}\\
    Adult & 0.917 & 0.915 & \colorbox[HTML]{E0EAF8}{0.919} & 0.881 & 0.907 & \colorbox[HTML]{A8B6E0}{0.931}\\
    Bank 32 nh & \colorbox[HTML]{A8B6E0}{0.887} & \colorbox[HTML]{A8B6E0}{0.887} & 0.881 & 0.859 & 0.876 & 0.874\\
    Covertype & 0.984 & \colorbox[HTML]{E0EAF8}{0.988} & 0.969 & 0.861 & \colorbox[HTML]{A8B6E0}{0.995} & 0.967\\
    Credit Card Fraud & \colorbox[HTML]{A8B6E0}{0.987} & 0.966 & 0.969 & 0.913 & 0.914 & \colorbox[HTML]{E0EAF8}{0.975}\\
    Delta Ailerons & 0.977 & 0.977 & 0.974 & 0.977 & \colorbox[HTML]{A8B6E0}{0.978} & 0.977\\
    Electricity & 0.901 & 0.929 & 0.894 & 0.928 & \colorbox[HTML]{E0EAF8}{0.97} & \colorbox[HTML]{A8B6E0}{0.973}\\
    Elevators & \colorbox[HTML]{A8B6E0}{0.951} & 0.948 & \colorbox[HTML]{E0EAF8}{0.95} & \colorbox[HTML]{E0EAF8}{0.95} & 0.913 & 0.943\\
    HPC Job Scheduling & 0.908 & 0.921 & 0.775 & 0.908 & \colorbox[HTML]{E0EAF8}{0.952} & \colorbox[HTML]{A8B6E0}{0.955}\\
    Fars & 0.956 & \colorbox[HTML]{E0EAF8}{0.958} & 0.954 & 0.955 & 0.949 & \colorbox[HTML]{A8B6E0}{0.962}\\
    1st Order Theorem Proving & 0.776 & 0.808 & 0.495 & 0.805 & \colorbox[HTML]{E0EAF8}{0.854} & \colorbox[HTML]{A8B6E0}{0.858}\\
    Helena & 0.875 & \colorbox[HTML]{E0EAF8}{0.889} & 0.884 & \colorbox[HTML]{A8B6E0}{0.897} & 0.855 & 0.875\\
    Heloc & \colorbox[HTML]{E0EAF8}{0.783} & \colorbox[HTML]{A8B6E0}{0.787} & 0.772 & \colorbox[HTML]{E0EAF8}{0.783} & 0.778 & 0.775\\
    Higgs & 0.762 & 0.785 & \colorbox[HTML]{A8B6E0}{0.804} & 0.774 & 0.793 & \colorbox[HTML]{E0EAF8}{0.797}\\
    Indian Pines & 0.984 & \colorbox[HTML]{A8B6E0}{0.992} & \colorbox[HTML]{E0EAF8}{0.99} & 0.973 & 0.979 & 0.987\\
    Jannis & 0.856 & 0.857 & \colorbox[HTML]{E0EAF8}{0.867} & 0.861 & 0.861 & \colorbox[HTML]{A8B6E0}{0.872}\\
    JM1 & \colorbox[HTML]{E0EAF8}{0.739} & 0.725 & 0.711 & 0.728 & \colorbox[HTML]{A8B6E0}{0.747} & 0.733\\
    LHC Identify Jets & \colorbox[HTML]{E0EAF8}{0.941} & \colorbox[HTML]{E0EAF8}{0.941} & \colorbox[HTML]{A8B6E0}{0.944} & 0.861 & 0.935 & \colorbox[HTML]{E0EAF8}{0.941}\\
    Madelon & \colorbox[HTML]{A8B6E0}{0.906} & 0.718 & 0.501 & 0.668 & 0.79 & \colorbox[HTML]{E0EAF8}{0.891}\\
    Magic Telescope & 0.907 & 0.921 & 0.927 & \colorbox[HTML]{E0EAF8}{0.929} & \colorbox[HTML]{A8B6E0}{0.934} & 0.928\\
    MC1 & \colorbox[HTML]{A8B6E0}{0.957} & 0.904 & 0.89 & 0.853 & 0.844 & \colorbox[HTML]{E0EAF8}{0.943}\\
    Mozilla4 & 0.954 & 0.961 & 0.971 & 0.963 & \colorbox[HTML]{E0EAF8}{0.988} & \colorbox[HTML]{A8B6E0}{0.99}\\
    Microaggregation2 & 0.778 & \colorbox[HTML]{A8B6E0}{0.792} & 0.752 & \colorbox[HTML]{E0EAF8}{0.782} & 0.768 & 0.781\\
    Numerai28.6 & \colorbox[HTML]{E0EAF8}{0.526} & \colorbox[HTML]{A8B6E0}{0.534} & 0.52 & 0.518 & 0.519 & 0.514\\
    Otto Group Product & 0.96 & 0.971 & 0.968 & 0.972 & \colorbox[HTML]{E0EAF8}{0.973} & \colorbox[HTML]{A8B6E0}{0.974}\\
    PC2 & \colorbox[HTML]{A8B6E0}{0.881} & 0.815 & \colorbox[HTML]{E0EAF8}{0.844} & 0.571 & 0.55 & 0.739\\
    Phonemes & 0.922 & \colorbox[HTML]{E0EAF8}{0.96} & 0.898 & 0.93 & \colorbox[HTML]{A8B6E0}{0.964} & 0.953\\
    Pollen & 0.492 & \colorbox[HTML]{E0EAF8}{0.508} & \colorbox[HTML]{A8B6E0}{0.509} & 0.496 & 0.489 & 0.467\\
    Satellite & \colorbox[HTML]{A8B6E0}{0.998} & 0.993 & 0.911 & 0.992 & \colorbox[HTML]{A8B6E0}{0.998} & 0.992\\
    Scene & \colorbox[HTML]{A8B6E0}{0.994} & 0.989 & 0.986 & \colorbox[HTML]{E0EAF8}{0.992} & 0.983 & 0.982\\
    Speed Dating & \colorbox[HTML]{E0EAF8}{0.853} & 0.835 & 0.797 & 0.822 & 0.845 & \colorbox[HTML]{A8B6E0}{0.86}\\ 
    Telco Customer Churn & \colorbox[HTML]{A8B6E0}{0.858} & \colorbox[HTML]{E0EAF8}{0.845} & 0.841 & 0.783 & 0.84 & 0.843\\
    Vehicle sensIT & \colorbox[HTML]{A8B6E0}{0.918} & \colorbox[HTML]{A8B6E0}{0.918} & 0.917 & 0.914 & 0.912 & 0.916\\ 
    Waveform-5000 & \colorbox[HTML]{A8B6E0}{0.965} & 0.962 & 0.933 & \colorbox[HTML]{A8B6E0}{0.965} & 0.959 & 0.957\\ \bottomrule
\end{tabular}
\end{adjustbox}
\label{table:experiments}
\vspace{4 mm}
\end{table*}

\section{Concluding Remarks}
\label{CR}

We have proposed IGNNet, an algorithm for tabular data classification, which exploits graph neural networks to produce transparent models. In contrast to post-hoc explanation techniques, IGNNet does not approximate or require costly computations, but provides the explanation while computing the prediction, and where the explanation prescribes exactly how the prediction is computed.

We have presented results from a large-scale empirical investigation, in which IGNNet was evaluated with respect to explainability and predictive performance. IGNNet was shown to generate explanations with feature scores aligned with the Shapley values without further computational cost. IGNNet was also shown to achieve a similar predictive performance as XGBoost, Random Forests, TabNet, and MLP, which are all well-known for their ability to generate high-performing models. 

One direction for future research is to explore approaches to model feature interactions in the adjacency matrix that go beyond linear correlations. Understanding how such non-linear interactions between features may impact the model's interpretability could be an intriguing area of exploration. A second direction is to investigate alternative encoders for categorical features rather than relying on one-hot encoding. It would also be interesting to extend IGNNet to handle non-tabular datasets, including images and text, which would require entirely different approaches to representing each data point as a graph.
Another important direction for future work is to use IGNNet for studying possible adversarial attacks on a predictive model. Finally, an important direction would be to complement the empirical evaluation with user-grounded evaluations, e.g., measuring to what extent certain tasks could be more effectively and efficiently solved when the users are provided with a transparent model that shows how the prediction has been computed from the input.



\begin{ack}
This work was partially supported by the Wallenberg AI, Autonomous Systems and Software Program (WASP) funded by the Knut and Alice Wallenberg Foundation.
\end{ack}



\bibliography{bibfile}

\appendix

\section{Illustration of Explanations}\label{case_study}

\begin{figure}[h]
  \centering
  \subfloat[The original data point.]{\includegraphics[width=0.49\textwidth]{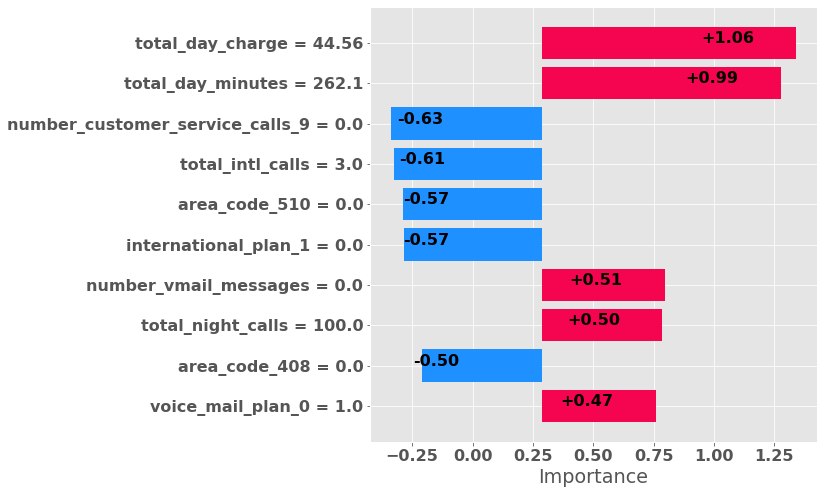}\label{fig:churn_original}}
  \hfill
  \vspace{5 mm}
  \subfloat[The data point with a modified total day charge value.]{\includegraphics[width=0.49\textwidth]{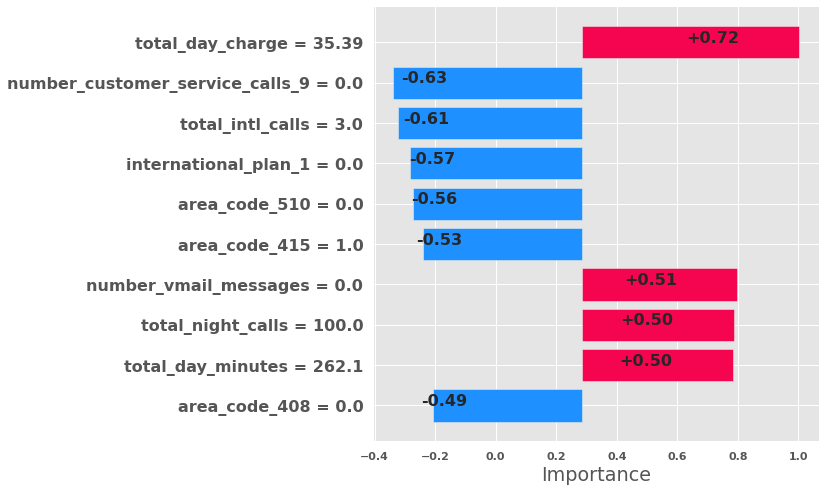}\label{fig:churn_mod}}
  \vspace{5 mm}
  \caption{The explanation of a single prediction on Churn dataset.}
  \vspace{5 mm}
\end{figure}

\begin{figure}[h]
    \centering
    \includegraphics[width=0.49\textwidth]{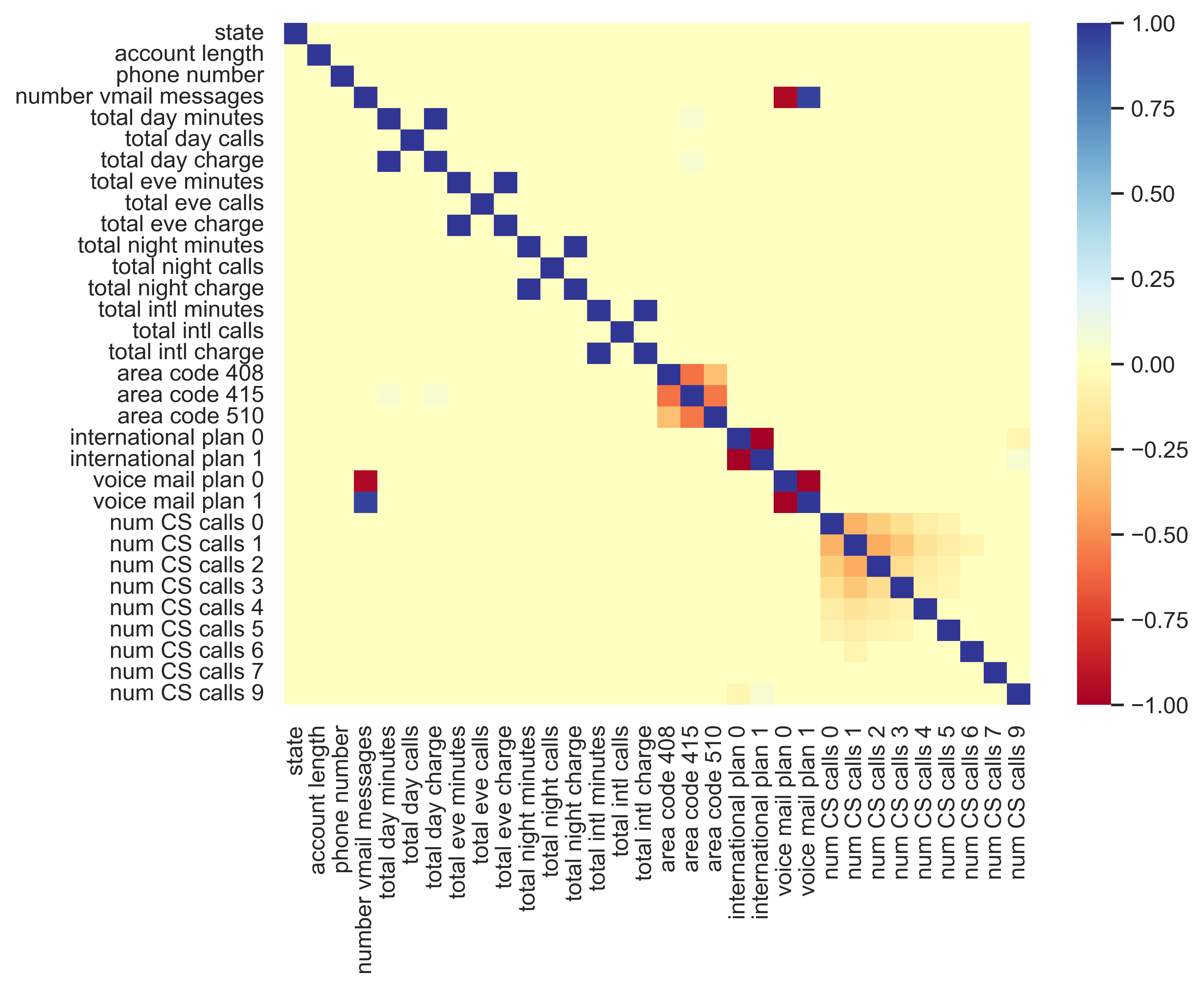}
    \vspace{1 mm}
    \caption{The correlation matrix of the features of the Churn dataset.}
    \vspace{5 mm}
    \label{fig:churn_corr_mat}
\end{figure}

In this section, we illustrate the computed feature scores by IGNNet on one example from the Churn dataset and show how the explanation can reflect the interaction between features.
The data point, from the Churn dataset, has a  positive prediction with a narrow margin (0.565). Consequently, the reduction of the top positively important feature (total day charge), as illustrated in Fig.~\ref{fig:churn_original}, may be enough to obtain a negative prediction. The total day charge has a maximum value of 59.76, a minimum of 0.44, a mean of 30.64, and a 9.17 standard deviation in the training set. However, the total day charge is highly correlated with the total day minutes by more than 0.99, as shown in Fig.~\ref{fig:churn_corr_mat}. The high correlation effect is obvious in the outcome when the total day charge value is reduced by one standard deviation, from 44.56 to 35.39, as the scores of both total day charge and total day minutes drop from 1.06 and 0.99 to 0.72 and 0.5, respectively, as shown in Fig.~\ref{fig:churn_mod}. Moreover, the sum of feature scores and bias falls below 0.5 after the sigmoid function, resulting in a negative prediction with a predicted value of 0.297. 

\clearpage
\section{Ablation Study}\label{ablation}

In this section, we conduct three sets of experiments to determine the effects of different hyperparameter choices on predictive performance. In the first set, we study the effects of changing the weight on the edge from a feature to itself in the adjacency matrix (self-loop). In the second set, we inspect the effect of enforcing a threshold on the correlation value between two features to be included as an edge weight. Finally, the third set of experiments determines the effect of using a different number of message-passing layers. 

\subsection{Weight on Self-Loop}

\begin{figure*}
    \centering
    \includegraphics[width=1.0\textwidth]{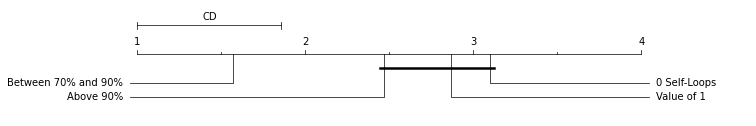}
    \vspace{1 mm}
    \caption{The average rank of IGNNet in the default settings and other IGNNet models with different values for the self-loops on the 30 datasets. The ranking has been done for the predictive performance measured in the AUC.}
    \vspace{5 mm}
    \label{fig:self-loops-rank}
\end{figure*}

In the experimental setup, we evaluated IGNNet using a value that forms 70\% up to 90\% of the weighted summation on the self-loop. In this set of experiments, we evaluate IGNNet using self-loops that form more than 90\% of the weighted summation, then we test with a fixed value of 1 for the self-loop, and finally, we evaluate with 0 self-loops. The detailed results are available in table \ref{table:ablation}. The null hypothesis that there is no difference in the predictive performance between IGNNet in the default settings, and IGNNet with the three new variations of the self-loops, has been tested using the Friedman test followed by the pairwise Nemenyi tests. The null hypothesis can be rejected at 0.05 level for the difference between the default settings and the tested variations of self-loops, as the default settings showed significantly better performance. The result of the post hoc tests are summarized in Fig.~\ref{fig:self-loops-rank}.

\subsection{Threshold on Correlation Values}

In the experimental setup, we set a threshold for a correlation value to be added as an edge between two features, which is 0.2 in general and 0.05 in the case of small correlation values. In this set of experiments, we compare the performance of IGNNet with the abovementioned thresholds and without any thresholds. To test the null hypothesis that there is no difference in the predictive performance, as measured by AUC, between the two variations, and since we compare two models, the Wilcoxon signed-rank test \cite{wilcoxon1945individual} is used here. The null hypothesis may be rejected at the 0.05 level, showing that setting a threshold for the edge weight helps with improving the predictive performance of IGNNet. The results are available in Table~\ref{table:ablation}.

\subsection{The Number of Message-Passing Layers}

In this final set of experiments, we investigate the effect of reducing the number of the message-passing layers and the total number of parameters in the IGNNet on the predictive performance. We compare IGNNet (with the default architecture) to a reduced version of three message-passing layers and another version with just one message-passing layer. For the three layers version, we use the architecture of IGNNet as mentioned in the experimental setup, but we keep the first three message-passing layers up to the first batch normalization, then we remove the following three message-passing layers and their linear transformations before the final fully-connected feedforward network. The model with one message-passing layer has only the first layer with the subsequent transformations, then all the subsequent layers are removed up until the fully-connected feedforward network is.

\begin{figure*}
    \centering
    \includegraphics[width=1.0\textwidth]{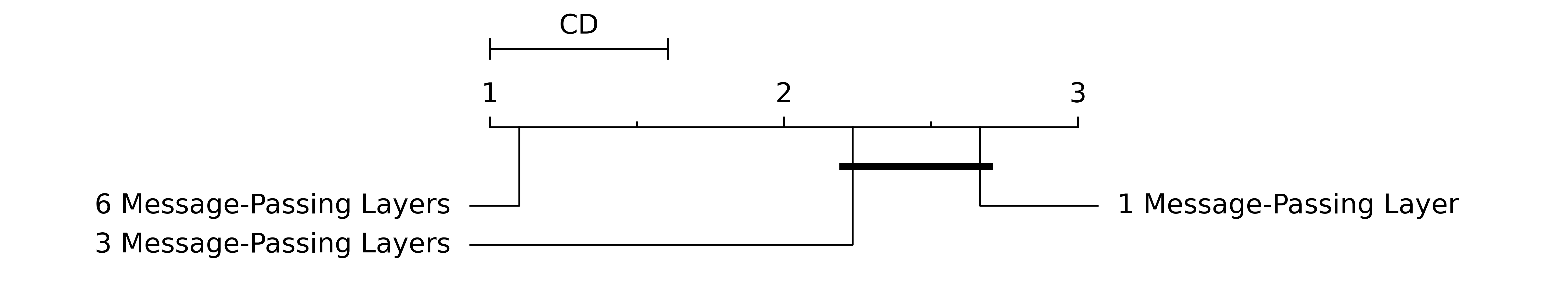}
    \vspace{1 mm}
    \caption{The average rank of IGNNet with 6, 3, and 1 message-passing layers on the 30 datasets. The ranking has been done for the predictive performance measured in the AUC.}
    \vspace{5 mm}
    \label{fig:num-MP-rank}
\end{figure*}

Similar to what has been done in the previous subsections, we test the null hypothesis of no difference in the predictive performance between the three architectures using the Friedman test, followed by pairwise post hoc Nemenyi tests. The first test rejects the null hypothesis, and the result of the Nemenyi tests are outlined in Fig.~\ref{fig:num-MP-rank}. The results in Table~\ref{table:ablation} demonstrate that the predictive performance generally improves with more layers of message-passing and linear transformations.

\begin{table*}
\caption{The results of the ablation study that compares the predictive performance of IGNNet in the default settings to IGNNet with different weights on the self-loop, IGNNet without a threshold on the correlation value to be included as an edge in the adjacency matrix, and IGNNet with different numbers of message-passing layers. The best-performing model is \colorbox[HTML]{A8B6E0}{colored in blue}, and the second best-performing is \colorbox[HTML]{E0EAF8}{colored in light blue}.}
\vspace{2 mm}
\centering
\begin{adjustbox}{angle=90, width=.8\textwidth}
\small
\begin{tabular}{lc|ccc|c|cc}
\toprule
\rowcolor[HTML]{EFEFEF} 
\multicolumn{1}{c}{\cellcolor[HTML]{EFEFEF}} & \multicolumn{1}{c|}{\cellcolor[HTML]{EFEFEF}} & \multicolumn{3}{c|}{\cellcolor[HTML]{EFEFEF}Self-Loop Experiments} & \multicolumn{1}{c|}{\cellcolor[HTML]{EFEFEF}} & \multicolumn{2}{c}{\cellcolor[HTML]{EFEFEF}Number of Layers Experiments} \\ \cline{3-5} \cline{7-8} 
\rowcolor[HTML]{EFEFEF} 
\multicolumn{1}{c}{\multirow{-2}{*}{\cellcolor[HTML]{EFEFEF}Dataset}} & \multicolumn{1}{c|}{\multirow{-2}{*}{\cellcolor[HTML]{EFEFEF}Default Settings}} & Above 90\% & Self-Loop Value = 1 & Self-Loop Value = 0 & \multicolumn{1}{c|}{\multirow{-2}{*}{\cellcolor[HTML]{EFEFEF}Without Correlation Threshold}} & 3 Layers & 1 Layer\\ \cmidrule(lr){1-8}
Abalone & \cellcolor[HTML]{A8B6E0}0.88 & \cellcolor[HTML]{A8B6E0}0.88 & 0.844 & 0.842 & \cellcolor[HTML]{A8B6E0}0.88 & 0.874 & 0.848\\ \cmidrule(lr){1-8}
Ada Prior & \cellcolor[HTML]{A8B6E0}0.905 & \cellcolor[HTML]{A8B6E0}0.905 & 0.891 & 0.88 & 0.896 & 0.904 & 0.903\\ \cmidrule(lr){1-8}
Adult & \cellcolor[HTML]{A8B6E0}0.917 & 0.914 & 0.916 & 0.88 & 0.916 & \cellcolor[HTML]{A8B6E0}0.917 & 0.911\\ \cmidrule(lr){1-8}
Bank 32 nh & \cellcolor[HTML]{E0EAF8}0.887 & \cellcolor[HTML]{A8B6E0}0.889 & 0.854 & 0.815 & 0.886 & 0.877 & 0.881\\ \cmidrule(lr){1-8}
Breast Cancer & \cellcolor[HTML]{A8B6E0}0.999 & \cellcolor[HTML]{A8B6E0}0.999 & 0.998 & 0.998 & \cellcolor[HTML]{A8B6E0}0.999 & 0.998 & 0.998\\ \cmidrule(lr){1-8}
Churn & 0.88 & 0.865 & 0.871 & \cellcolor[HTML]{E0EAF8}0.9 & \cellcolor[HTML]{A8B6E0}0.905 & 0.872 & 0.868\\ \cmidrule(lr){1-8}
Credit Card Fraud & \cellcolor[HTML]{A8B6E0}0.987 & \cellcolor[HTML]{E0EAF8}0.974 & 0.942 & 0.913 & 0.952 & 0.938 & 0.937\\ \cmidrule(lr){1-8}
Delta Ailerons & \cellcolor[HTML]{A8B6E0}0.977 & \cellcolor[HTML]{A8B6E0}0.977 & \cellcolor[HTML]{A8B6E0}0.977 & 0.976 & \cellcolor[HTML]{A8B6E0}0.977 & 0.976 & \cellcolor[HTML]{A8B6E0}0.977\\ \cmidrule(lr){1-8}
Delta Elevators & \cellcolor[HTML]{E0EAF8}0.951 & \cellcolor[HTML]{A8B6E0}0.952 & 0.946 & 0.946 & \cellcolor[HTML]{E0EAF8}0.951 & 0.95 & 0.949\\ \cmidrule(lr){1-8}
Electricity & \cellcolor[HTML]{A8B6E0}0.898 & 0.871 & \cellcolor[HTML]{E0EAF8}0.881 & 0.844 & 0.866 & 0.845 & 0.838\\ \cmidrule(lr){1-8}
Elevators & \cellcolor[HTML]{A8B6E0}0.951 & \cellcolor[HTML]{E0EAF8}0.946 & 0.923 & 0.928 & 0.931 & 0.896 & 0.847\\ \cmidrule(lr){1-8}
Higgs & \cellcolor[HTML]{E0EAF8}0.762 & \cellcolor[HTML]{A8B6E0}0.791 & 0.696 & 0.707 & 0.758 & 0.727 & 0.717\\ \cmidrule(lr){1-8}
JM1 & \cellcolor[HTML]{A8B6E0}0.739 & 0.729 & 0.702 & 0.71 & \cellcolor[HTML]{E0EAF8}0.734 & 0.721 & 0.723\\ \cmidrule(lr){1-8}
Madelon & \cellcolor[HTML]{E0EAF8}0.906 & 0.717 & \cellcolor[HTML]{A8B6E0}0.91 & 0.895 & 0.855 & 0.807 & 0.686\\ \cmidrule(lr){1-8}
Magic Telescope & 0.907 & \cellcolor[HTML]{A8B6E0}0.929 & 0.89 & 0.91 & \cellcolor[HTML]{E0EAF8}0.919 & 0.901 & 0.873\\ \cmidrule(lr){1-8}
Mozilla4 & 0.954 & 0.953 & 0.954 & \cellcolor[HTML]{A8B6E0}0.965 & \cellcolor[HTML]{E0EAF8}0.963 & 0.942 & 0.862\\ \cmidrule(lr){1-8}
MC1 & \cellcolor[HTML]{E0EAF8}0.957 & 0.909 & 0.953 & 0.935 & 0.933 & 0.904 & \cellcolor[HTML]{A8B6E0}0.961\\ \cmidrule(lr){1-8}
Numerai28.6 & \cellcolor[HTML]{A8B6E0}0.526 & 0.52 & 0.52 & 0.521 & 0.52 & \cellcolor[HTML]{E0EAF8}0.525 & \cellcolor[HTML]{E0EAF8}0.525\\ \cmidrule(lr){1-8}
PC2 & \cellcolor[HTML]{E0EAF8}0.881 & \cellcolor[HTML]{A8B6E0}0.883 & 0.865 & 0.859 & 0.827 & 0.849 & 0.875\\ \cmidrule(lr){1-8}
Phishing & 0.99 & 0.987 & 0.989 & \cellcolor[HTML]{A8B6E0}0.993 & \cellcolor[HTML]{E0EAF8}0.992 & 0.988 & 0.987\\ \cmidrule(lr){1-8}
Phonemes & 0.922 & 0.919 & \cellcolor[HTML]{E0EAF8}0.924 & 0.89 & \cellcolor[HTML]{A8B6E0}0.941 & 0.915 & 0.895\\ \cmidrule(lr){1-8}
Pollen & \cellcolor[HTML]{A8B6E0}0.525 & 0.491 & 0.516 & \cellcolor[HTML]{E0EAF8}0.524 & 0.483 & 0.489 & 0.498\\ \cmidrule(lr){1-8}
Satellite & \cellcolor[HTML]{A8B6E0}0.998 & \cellcolor[HTML]{E0EAF8}0.996 & 0.928 & 0.931 & 0.984 & \cellcolor[HTML]{E0EAF8}0.996 & 0.995\\ \cmidrule(lr){1-8}
Scene & \cellcolor[HTML]{A8B6E0}0.994 & 0.959 & 0.969 & 0.932 & \cellcolor[HTML]{E0EAF8}0.992 & 0.957 & 0.941\\ \cmidrule(lr){1-8}
Spambase & \cellcolor[HTML]{E0EAF8}0.984 & 0.98 & 0.974 & 0.963 & 0.983 & \cellcolor[HTML]{A8B6E0}0.985 & 0.982\\ \cmidrule(lr){1-8}
Speed Dating & \cellcolor[HTML]{A8B6E0}0.853 & \cellcolor[HTML]{E0EAF8}0.844 & 0.825 & 0.812 & 0.824 & 0.82 & 0.814\\ \cmidrule(lr){1-8}
Telco Customer Churn & \cellcolor[HTML]{E0EAF8}0.858 & \cellcolor[HTML]{A8B6E0}0.86 & 0.841 & 0.844 & \cellcolor[HTML]{E0EAF8}0.858 & 0.852 & 0.853\\ \cmidrule(lr){1-8}
Tic Tac Toe & 0.846 & 0.828 & 0.841 & \cellcolor[HTML]{E0EAF8}0.864 & \cellcolor[HTML]{A8B6E0}0.876 & 0.831 & 0.803\\ \cmidrule(lr){1-8}
Vehicle sensIT & \cellcolor[HTML]{A8B6E0}0.918 & \cellcolor[HTML]{E0EAF8}0.916 & 0.915 & 0.91 & 0.915 & 0.915 & 0.914\\ \cmidrule(lr){1-8}
Waveform-5000 & \cellcolor[HTML]{E0EAF8}0.965 & 0.962 & \cellcolor[HTML]{A8B6E0}0.966 & 0.964 & 0.961 & 0.96 & 0.96\\ \bottomrule

\end{tabular}
\end{adjustbox}
\label{table:ablation}
\vspace{4 mm}
\end{table*}

\section{Transparency Evaluation}

\begin{figure*}[ht]

\centering
\includegraphics[width=.34\textwidth]{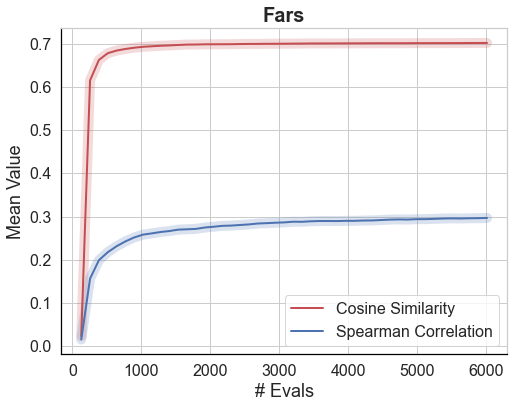}\hfill
\includegraphics[width=.33\textwidth]{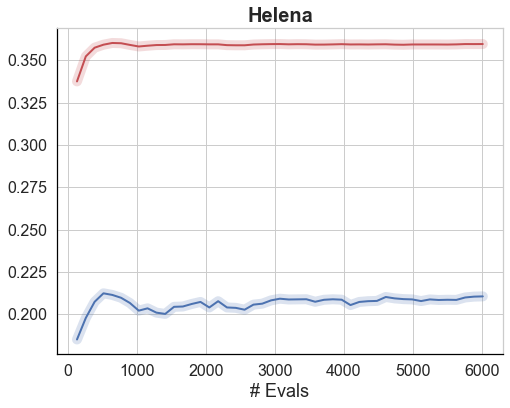}\hfill
\includegraphics[width=.32\textwidth]{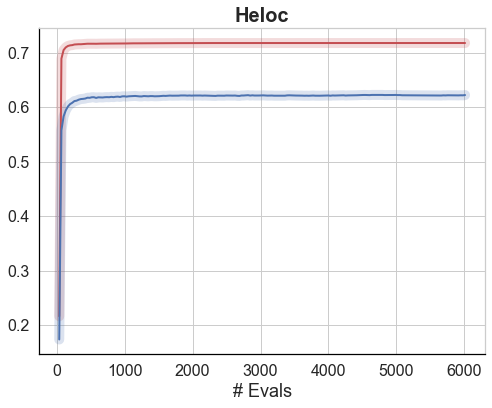}
\vspace{1 mm}
\includegraphics[width=.34\textwidth]{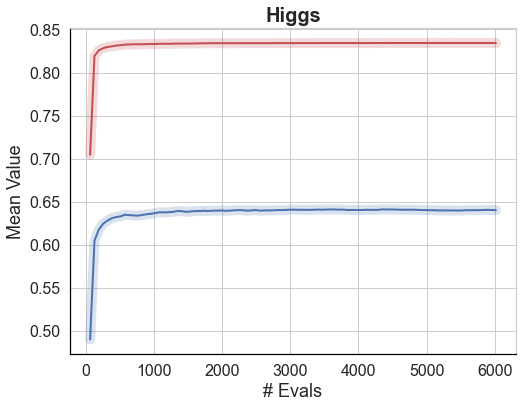}\hfill
\includegraphics[width=.33\textwidth]{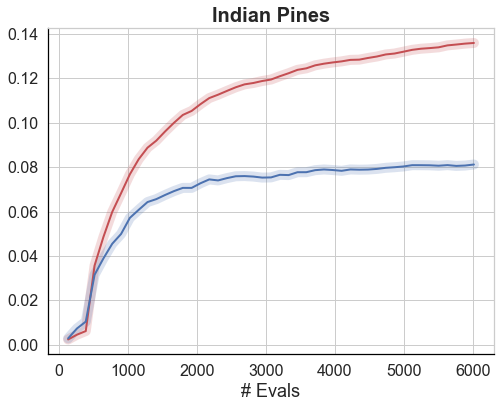}\hfill
\includegraphics[width=.32\textwidth]{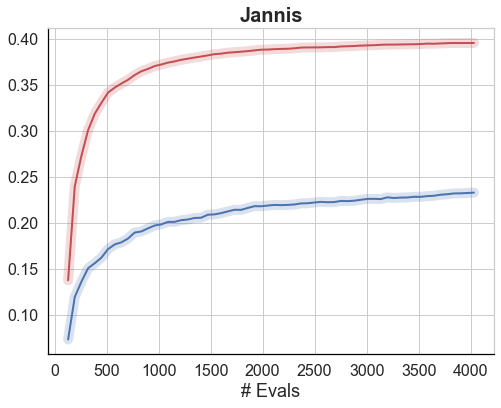}
\vspace{1 mm}
\includegraphics[width=.34\textwidth]{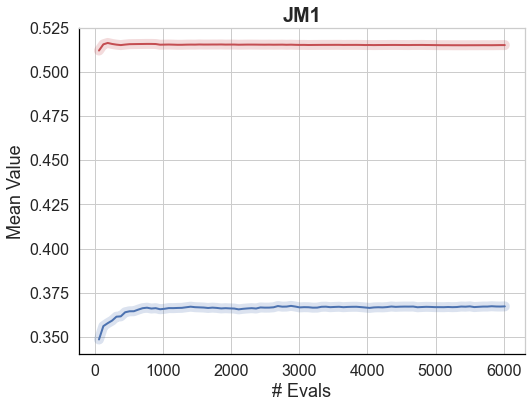}\hfill
\includegraphics[width=.33\textwidth]{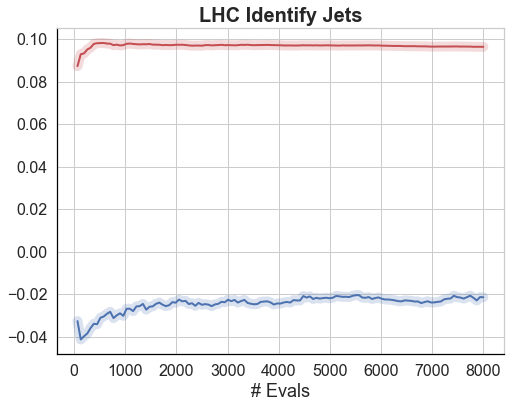}\hfill
\includegraphics[width=.32\textwidth]{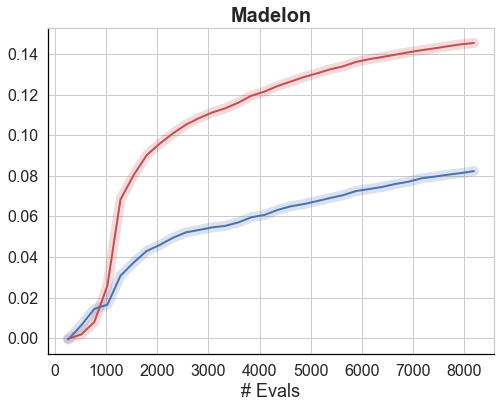}
\vspace{1 mm}
\includegraphics[width=.34\textwidth]{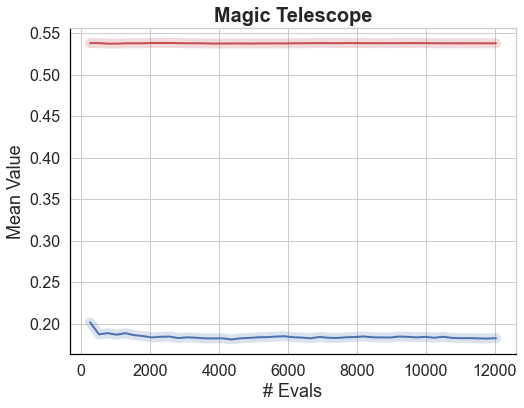}\hfill
\includegraphics[width=.33\textwidth]{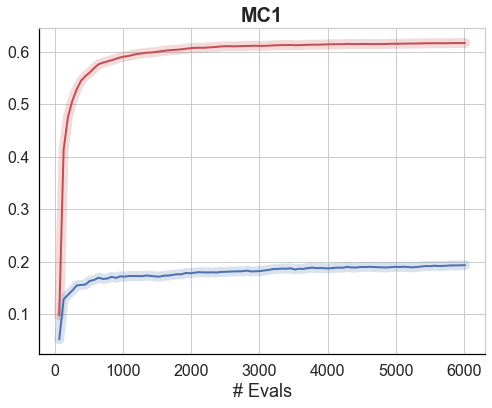}\hfill
\includegraphics[width=.32\textwidth]{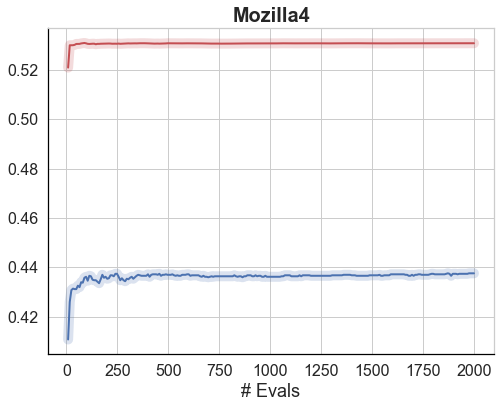}
\vspace{1 mm}
\caption{\textbf{Comparison of KernelSHAP's approximations and the importance scores obtained from IGNNet.} We measure the similarity of KernelSHAP's approximations to the scores of IGNNet at each iteration of data sampling and evaluation of KernelSHAP. KernelSHAP exhibits improvement in approximating the scores derived from IGNNet with more data sampling.}
\label{fig:transparency2}

\end{figure*}

\begin{figure*}[ht]

\centering
\includegraphics[width=.34\textwidth]{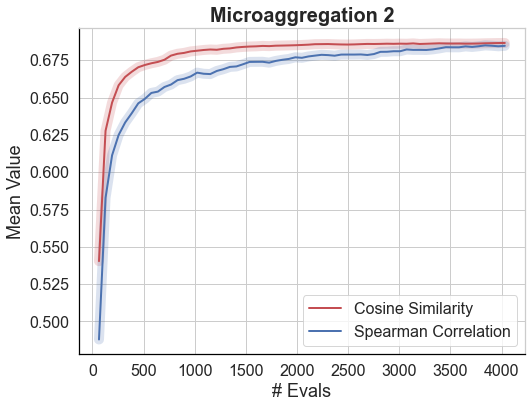}\hfill
\includegraphics[width=.33\textwidth]{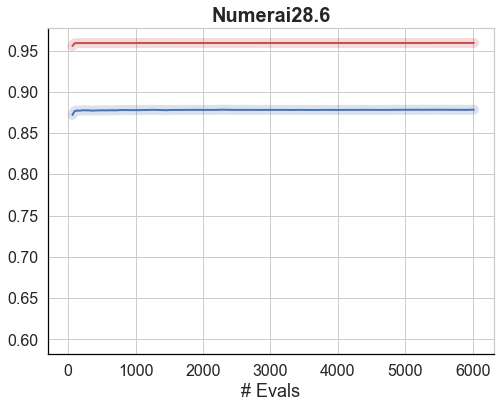}\hfill
\includegraphics[width=.32\textwidth]{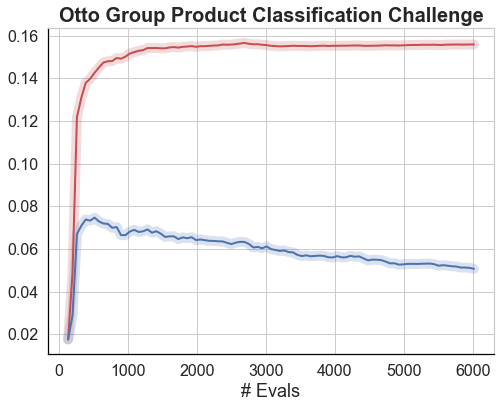}
\vspace{1 mm}
\includegraphics[width=.34\textwidth]{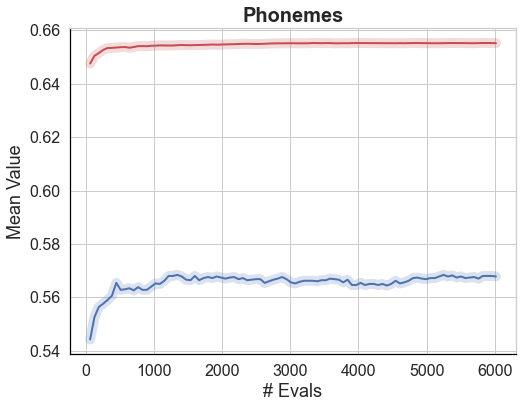}\hfill
\includegraphics[width=.33\textwidth]{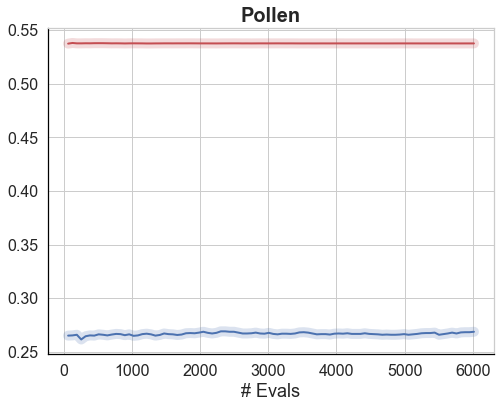}\hfill
\includegraphics[width=.32\textwidth]{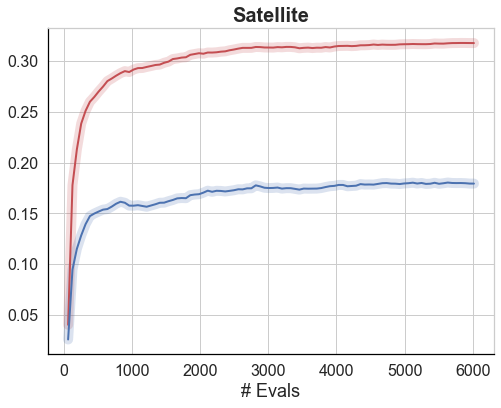}
\vspace{1 mm}
\includegraphics[width=.34\textwidth]{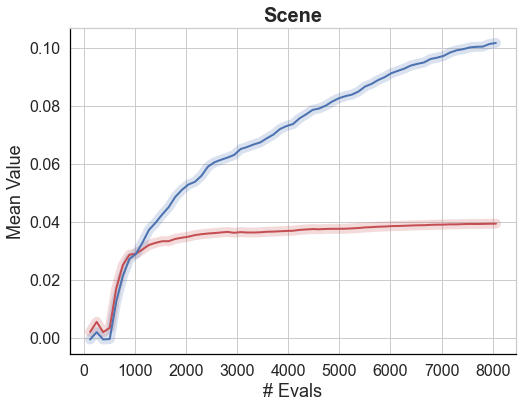}\hfill
\includegraphics[width=.33\textwidth]{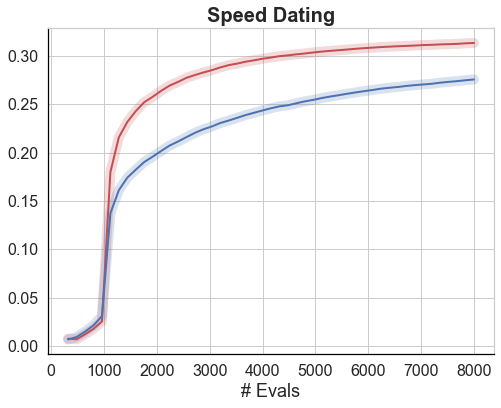}\hfill
\includegraphics[width=.32\textwidth]{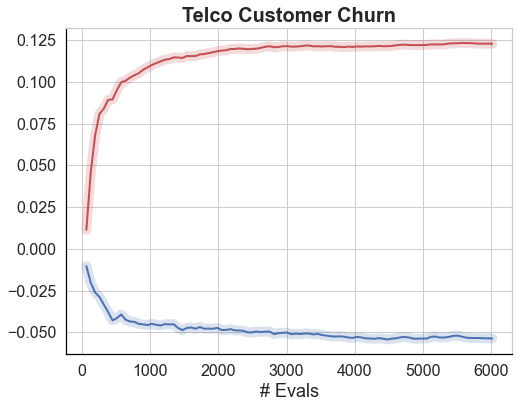}
\vspace{1 mm}
\includegraphics[width=.34\textwidth]{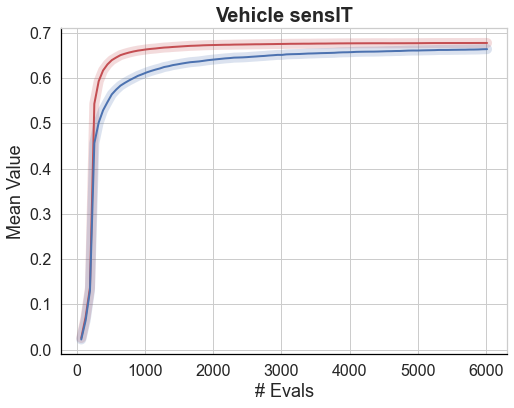}
\includegraphics[width=.34\textwidth]{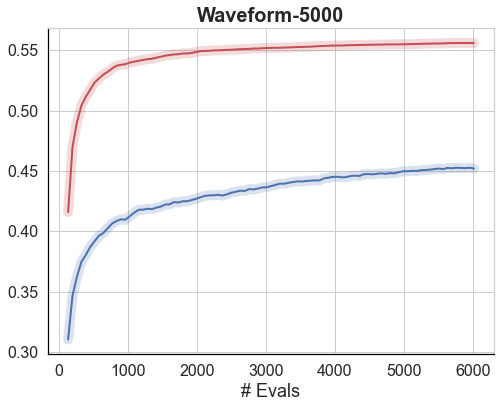}

\vspace{1 mm}
\caption{\textbf{Comparison of KernelSHAP's approximations and the importance scores obtained from IGNNet.} We measure the similarity of KernelSHAP's approximations to the scores of IGNNet at each iteration of data sampling and evaluation of KernelSHAP. KernelSHAP exhibits improvement in approximating the scores derived from IGNNet with more data sampling.}
\label{fig:transparency3}

\end{figure*}

In this section, we demonstrate the detailed results of the explanations evaluation experiment using 35 datasets. The results show a general trend across the 35 datasets where the explanations obtained using KernelSHAP converge to more similar values to the feature scores produced by IGNNet, given more sampled data. The results are displayed in Figure \ref{fig:transparency}, Figure \ref{fig:transparency2}, and Figure \ref{fig:transparency3}.

\begin{figure*}[ht]

\centering
\includegraphics[width=.34\textwidth]{figures/expl-bank-plt.png}\hfill
\includegraphics[width=.32\textwidth]{figures/expl-1st-order-plt.png}\hfill
\includegraphics[width=.32\textwidth]{figures/expl-pc2.png}
\vspace{1 mm}
\includegraphics[width=.34\textwidth]{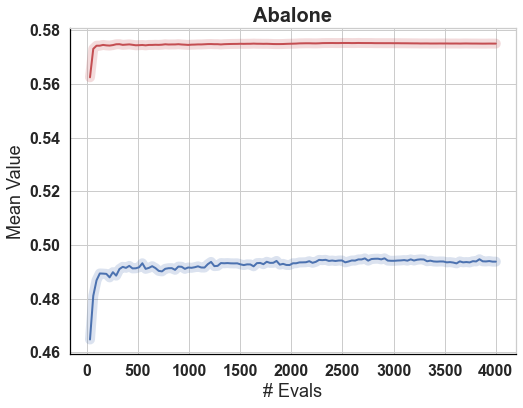}\hfill
\includegraphics[width=.32\textwidth]{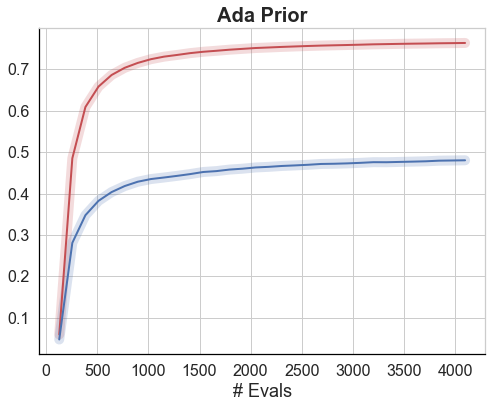}\hfill
\includegraphics[width=.32\textwidth]{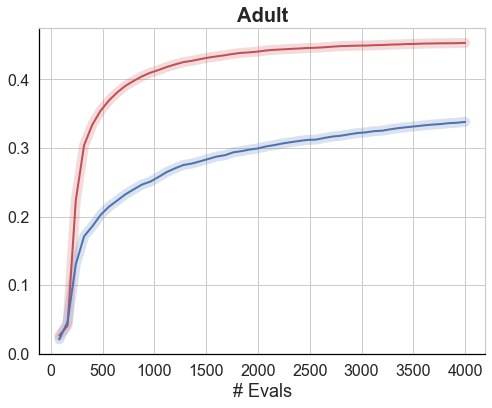}
\vspace{1 mm}
\includegraphics[width=.34\textwidth]{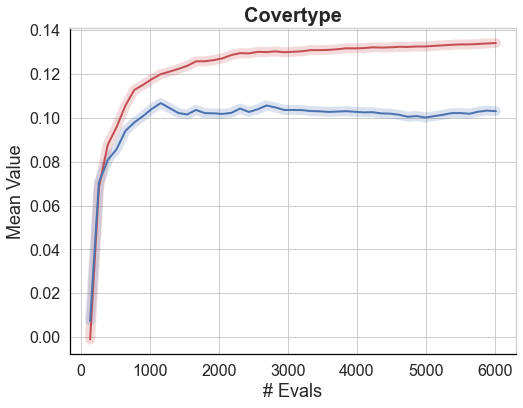}\hfill
\includegraphics[width=.32\textwidth]{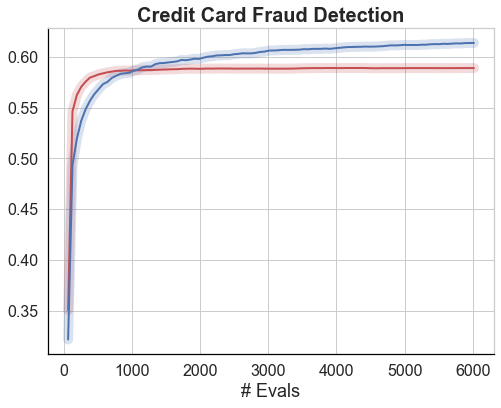}\hfill
\includegraphics[width=.32\textwidth]{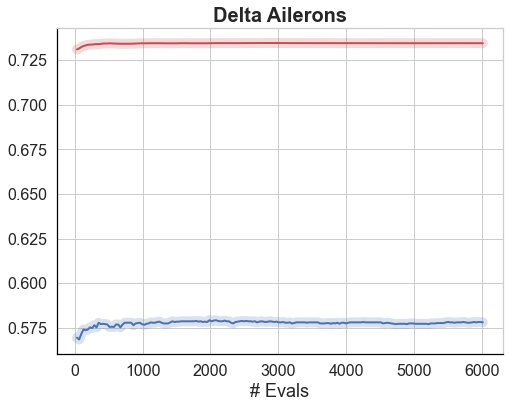}
\vspace{1 mm}
\includegraphics[width=.34\textwidth]{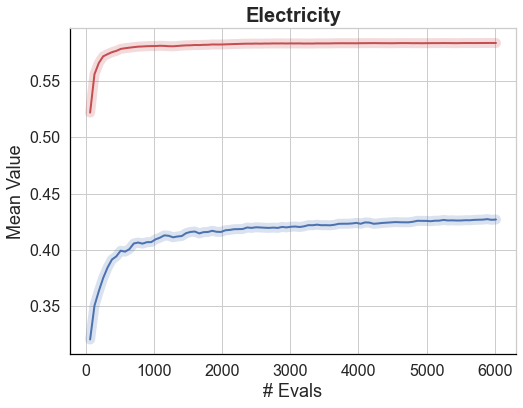}\hfill
\includegraphics[width=.32\textwidth]{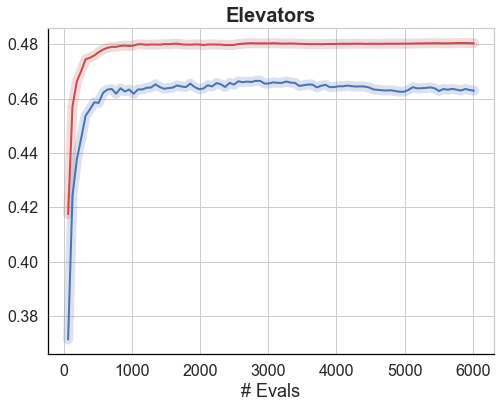}\hfill
\includegraphics[width=.32\textwidth]{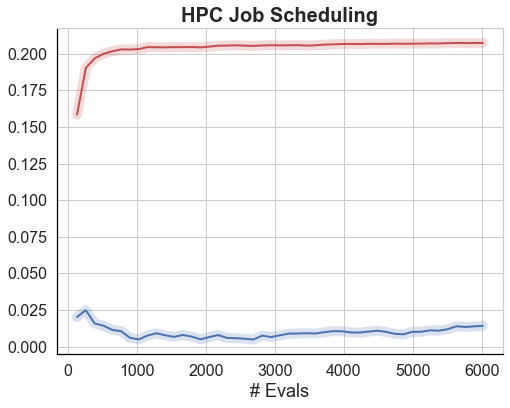}

\vspace{1 mm}
\caption{\textbf{Comparison of KernelSHAP's approximations and the importance scores obtained from IGNNet.} We measure the similarity of KernelSHAP's approximations to the scores of IGNNet at each iteration of data sampling and evaluation of KernelSHAP. KernelSHAP exhibits improvement in approximating the scores derived from IGNNet with more data sampling.}
\label{fig:transparency}

\end{figure*}

\section{Information about the Used Datasets and Specifications of the Hardware}
This subsection provides a summary of the datasets utilized in the experiments. In Table~\ref{table:datasets}, we provide information about the used datasets, including the number of classes (Num of Classes), the number of features	(Num of Features), the size of the dataset, the size of the training, validation, and test splits, as well as, the used correlation threshold of each dataset, the weight on the self-loop (Self-Loop Wt.) and finally the ID of each dataset on OpenML. 

The experiments have been performed in a Python environment on an Intel(R) Core(TM) i9-10885H CPU @ 2.40GHz system with 64.0 GB of RAM, and the GPUs are NVIDIA® GeForce® GTX 1650 Ti with 4 GB GDDR6, and NVIDIA® GeForce® GTX 1080 Ti with 8 GB. All the software and package dependencies are documented with the source code.

\begin{table*}
\caption{The dataset information.}
\vspace{2 mm}
\centering
\begin{adjustbox}{angle=90, width=.7\textwidth}
\small
\begin{tabular}{l c c c c c c c c c c}
    \toprule
    \rowcolor[HTML]{EFEFEF} 

    \multicolumn{1}{c}{{\cellcolor[HTML]{EFEFEF}Dataset}} & \multicolumn{1}{l}{\cellcolor[HTML]{EFEFEF}Classes} & Features     & Dataset Size & Training Set    & Dev. Set   & Test Set & Corr. Threshold & Self-Loop Wt. & Training Epochs & OpenML ID    \\
    \cmidrule(lr){1-11}
    Abalone & 2 & 8 & 4177 & 2506 & 836 & 835 & 0.2 & 20 & 220 & 720\\ 
    Ada Prior & 2 & 14 & 4562 & 2737 & 913 & 912 & 0.2 & 4 & 216 & 1037\\
    Adult & 2 & 14 & 48842 & 43957 & 2443 & 2442 & 0.2 & 4 & 297 & 1590\\
    Bank 32 nh & 2 & 32 & 8192 & 5734 & 1229 & 1229 & 0.2 & 2 & 540 & 833\\
    Covertype & 7 & 54 & 581012 & 524362 & 27599 & 29051 & 0.2 & 10 & 300 & 1596\\
    Credit Card Fraud & 2 & 30 & 284807 & 270566 & 7121 & 7120 & 0.2 & 3 & 29 & 42175\\
    Delta Ailerons & 2 & 5 & 7129 & 3564 & 1783 & 1782 & 0.2 & 2 & 380 & 803\\
    Electricity & 2 & 8 & 45312 & 36249 & 4532 & 4531 & 0.2 & 3 & 396 & 151\\
    Elevators & 2 & 18 & 16599 & 11619 & 2490 & 2490 & 0.2 & 4 & 353 & 846\\
    HPC Job Scheduling & 4 & 7 & 4331 & 2598 & 867 & 866 & 0.05 & 10 & 435 & 43925\\
    Fars & 8 & 29 & 100968 & 80774 & 10097 & 10097 & 0.05 & 10 & 301 & 40672\\
    1st Order Theorem Prov. & 6 & 51 & 6118 & 3915 & 979 & 1224 & 0.2 & 30 & 848 & 1475\\
    Helena & 100 & 27 & 65196 & 41724 & 10432 & 13040 & 0.2 & 10 & 550 & 41169\\
    Heloc & 2 & 22 & 10000 & 7500 & 1250 & 1250 & 0.2 & 20 & 234 & 45023\\
    Higgs & 2 & 28 & 98050 & 88245 & 4903 & 4902 & 0.05 & 4 & 394 & 23512\\
    Indian Pines & 8 & 220 & 9144 & 5852 & 1463 & 1829 & 0.2 & 400 & 394 & 41972\\
    Jannis & 4 & 54 & 83733 & 53588 & 13398 & 16747 & 0.05 & 20 & 300 & 41168\\
    JM1 & 2 & 21 & 10885 & 8708 & 1089 & 1088 & 0.2 & 50 & 187 & 1053\\
    LHC Identify Jets & 5 & 16 & 830000 & 749075 & 39425 & 41500 & 0.2 & 10 & 300 & 42468\\
    Madelon & 2 & 500 & 2600 & 1560 & 520 & 520 & 0.05 & 4 & 199 & 1485\\
    Magic Telescope & 2 & 10 & 19020 & 15216 & 1902 & 1902 & 0.2 & 4 & 397 & 1120\\
    MC1 & 2 & 38 & 9466 & 7478 & 994 & 994 & 0.2 & 80 & 198 & 1056\\
    Mozilla4 & 2 & 5 & 15545 & 12436 & 1555 & 1554 & 0.2 & 2 & 280 & 1046\\
    Microaggregation2 & 5 & 20 & 20000 & 12800 & 3200 & 4000 & 0.2 & 15 & 599 & 41671\\
    Numerai28.6 & 2 & 21 & 96320 & 86688 & 4816 & 4816 & 0.2 & 20 & 36 & 23517\\
    Otto Group Product & 9 & 93 & 61878 & 39601 & 9901 & 12376 & 0.05 & 30 & 270 & 45548\\
    PC2 & 2 & 36 & 5589 & 3353 & 1118 & 1118 & 0.2 & 60 & 37 & 1069\\
    Phonemes & 2 & 5 & 5404 & 3782 & 811 & 811 & 0.2 & 1 & 800 & 1489\\
    Pollen & 2 & 5 & 3848 & 2308 & 770 & 770 & 0.2 & 4 & 351 & 871\\
    Satellite & 2 & 36 & 5100 & 2805 & 1148 & 1147 & 0.2 & 60 & 287 & 40900\\
    Scene & 2 & 299 & 2407 & 1203 & 602 & 602 & 0.2 & 40 & 86 & 312\\
    Speed Dating & 2 & 120 & 8378 & 5864 & 1257 & 1257 & 0.2 & 10 & 58 & 40536\\
    Telco Customer Churn & 2 & 19 & 7043 & 4930 & 1057 & 1056 & 0.2 & 3 & 116 & 42178\\
    Vehicle sensIT & 2 & 100 & 98528 & 88675 & 4927 & 4926 & 0.2 & 10 & 172 & 357\\
    waveform-5000 & 2 & 40 & 5000 & 3000 & 1000 & 1000 & 0.2 & 4 & 97 & 979\\ \bottomrule
\end{tabular}
\end{adjustbox}
\label{table:datasets}
\vspace{4 mm}
\end{table*}

\section{Inference Time}

The inference time has been reported using an Intel(R) Core(TM) i9-10885H CPU @ 2.40GHz system with 64.0 GB of RAM, and the GPU is NVIDIA® GeForce® GTX 1650 Ti with 4 GB GDDR6. The time is reported as the average over 100 different data objects in milliseconds. The results are shown in Table~\ref{table:time}. Moreover, Table~\ref{table:time} shows the original number of features (Features), the subsequent number of nodes after the binarization of categorical features using one-hot-encoding (OHE Nodes), and the number of non-zero edges in the graph.

\begin{table*}
\caption{The inference time across different datasets.}
\vspace{2 mm}
\centering
\begin{adjustbox}{width=.8\textwidth}
\small
\begin{tabular}{l c c c c c}
    \toprule
    \rowcolor[HTML]{EFEFEF} 

    \multicolumn{1}{c}{{\cellcolor[HTML]{EFEFEF}Dataset}} & \multicolumn{1}{l}{\cellcolor[HTML]{EFEFEF}Classes} & Features     & OHE Nodes   & Non-Zero Edges     & Inference Time (ms) \\
    \cmidrule(lr){1-6}
    Abalone & 2 & 8 & 10 & 100 & 3.74 $\pm$ 0.77\\ 
    Ada Prior & 2 & 14 & 100 & 284 & 4.17 $\pm$ 0.7 \\
    Adult & 2 & 14 & 105 & 283 & 4.11 $\pm$ 0.69\\
    Bank 32 nh & 2 & 32 & 32 & 38 & 4.06 $\pm$ 0.72\\
    Covertype & 7 & 54 & 98 & 488 & 9.03 $\pm$ 1.35\\
    Credit Card Fraud & 2 & 30 & 30 & 50 & 3.93 $\pm$ 0.68\\
    Delta Ailerons & 2 & 5 & 5 & 11 & 3.91 $\pm$ 0.67\\
    Electricity & 2 & 8 & 14 & 38 & 3.87 $\pm$ 0.73\\
    Elevators & 2 & 18 & 18 & 112 & 4.31 $\pm$ 0.55\\
    HPC Job Scheduling & 4 & 7 & 26 & 258 & 8.48 $\pm$ 1.81\\
    Fars & 8 & 29 & 107 & 2295 & 8.18 $\pm$ 1.11\\
    1st Order Theorem Prov. & 6 & 51 & 51 & 1251 & 8.34 $\pm$ 0.92\\
    Helena & 100 & 27 & 27 & 293 & 4.61 $\pm$ 0.57\\
    Heloc & 2 & 22 & 22 & 370 & 5.48 $\pm$ 0.51\\
    Higgs & 2 & 28 & 28 & 244 & 6.58 $\pm$ 0.6\\
    Indian Pines & 8 & 220 & 220 & 40948 & 10.7 $\pm$ 1.53\\
    Jannis & 4 & 54 & 54 & 542 & 4.15 $\pm$ 0.7\\
    JM1 & 2 & 21 & 21 & 415 & 9.92 $\pm$ 0.79\\
    LHC Identify Jets & 5 & 16 & 16 & 184 & 3.85 $\pm$ 0.76\\
    Madelon & 2 & 500 & 500 & 12566 & 10.42 $\pm$ 2.02\\
    Magic Telescope & 2 & 10 & 10 & 54 & 5.88 $\pm$ 0.53\\
    MC1 & 2 & 38 & 38 & 1142 & 4.39 $\pm$ 0.71\\
    Mozilla4 & 2 & 5 & 5 & 7 & 4.03 $\pm$ 0.55\\
    Microaggregation2 & 5 & 20 & 20 & 138 & 3.82 $\pm$ 0.7\\
    Numerai28.6 & 2 & 21 & 21 & 231 & 4.68 $\pm$ 0.618\\
    Otto Group Product & 9 & 93 & 93 & 3581 & 4.06 $\pm$ 0.82\\
    PC2 & 2 & 36 & 36 & 1040 & 4.37 $\pm$ 0.73\\
    Phonemes & 2 & 5 & 5 & 9 & 4.31 $\pm$ 0.56\\
    Pollen & 2 & 5 & 5 & 15 & 4.31 $\pm$ 0.67\\
    Satellite & 2 & 36 & 36 & 1254 & 4.84 $\pm$ 0.44\\
    Scene & 2 & 299 & 304 & 19178 & 8.61 $\pm$ 1.27\\
    Speed Dating & 2 & 120 & 500 & 3454 & 7.12 $\pm$ 1.16\\
    Telco Customer Churn & 2 & 19 & 45 & 831 & 5.52 $\pm$ 0.65\\
    Vehicle sensIT & 2 & 100 & 100 & 1278 & 4.66 $\pm$ 0.83\\
    waveform-5000 & 2 & 40 & 40 & 306 & 4.62 $\pm$ 0.65\\ \bottomrule
\end{tabular}
\end{adjustbox}
\label{table:time}
\vspace{4 mm}
\end{table*}

\end{document}